\documentclass{article}

\usepackage{microtype}
\usepackage{graphicx}
\usepackage{subfigure}
\usepackage{booktabs} % for professional tables

\usepackage{hyperref}

\usepackage[ruled, linesnumbered]{algorithm2e}

 \usepackage[accepted]{icml2024}

\usepackage{amsmath}
\usepackage{amssymb}
\usepackage{mathtools}
\usepackage{amsthm}

\usepackage[capitalize,noabbrev]{cleveref}

\theoremstyle{plain}
\newtheorem{theorem}{Theorem}[section]

\newtheorem{lemma}[theorem]{Lemma}

\theoremstyle{definition}
\newtheorem{definition}[theorem]{Definition}

\theoremstyle{remark}

\usepackage[textsize=tiny]{todonotes}

\icmltitlerunning{Lorentz Structural Entropy Net}

\begin{document}

\twocolumn[
\icmltitle{\texttt{LSEnet}: Lorentz Structural Entropy Neural Network for\\Deep Graph Clustering}




\begin{icmlauthorlist}
\icmlauthor{Li Sun}{xxx}
\icmlauthor{Zhenhao Huang}{xxx}
\icmlauthor{Hao Peng}{yyy}
\icmlauthor{Yujie Wang}{xxx}
\icmlauthor{Chunyang Liu}{zzz}
\icmlauthor{Philip S. Yu}{aaa}
\end{icmlauthorlist}

\icmlaffiliation{xxx}{North China Electric Power University, Beijing 102206, China}
\icmlaffiliation{yyy}{Beihang University, Beijing 100191, China}
\icmlaffiliation{zzz}{Didi Chuxing, Beijing, China}
\icmlaffiliation{aaa}{University of Illinois at Chicago, IL, USA}

\icmlcorrespondingauthor{Li Sun}{ccesunli@ncepu.edu}
%\icmlcorrespondingauthor{Firstname2 Lastname2}{first2.last2@www.uk}

\icmlkeywords{Machine Learning, ICML}

\vskip 0.3in
]



\printAffiliationsAndNotice{}  % leave blank if no need to mention equal contribution

\begin{abstract}
Graph clustering is a fundamental problem in machine learning.
Deep learning methods achieve the state-of-the-art results in recent years, but they still cannot work without predefined cluster numbers.
Such limitation motivates us to pose a more challenging problem of \emph{graph clustering with unknown cluster number}.
We propose to address this problem from a fresh perspective of graph information theory (i.e., structural information).
In the literature, structural information has not yet been introduced to deep clustering, and its classic definition falls short of discrete formulation and modeling node features. 
In this work, we first formulate a differentiable structural information (DSI) in the continuous realm, accompanied by several theoretical results. 
% Minimizing DSI, the optimal partitioning tree is constructed where densely connected nodes in the graph tend to have the same assignment, revealing the cluster structure. 
By minimizing DSI, we construct the optimal partitioning tree where densely connected nodes in the graph tend to have the same assignment, revealing the cluster structure. 
DSI is also theoretically presented as a new graph clustering objective, not requiring the predefined cluster number. 
Furthermore, we design a neural \texttt{LSEnet} in the Lorentz model of hyperbolic space, where we integrate node features to structural information via manifold-valued graph convolution.
% DSI is shown to be a new objective of graph clus- tering, not requiring the predefined cluster number. Then, in the Lorentz model of hyperbolic space, we de- sign a neural LSEnet that further integrates node features via graph neural nets. LSEnet learns the optimal partitioning tree of a graph by mini- mizing DSI, and discovers the node clusters cor- respondingly. 
Extensive empirical results on real graphs show the superiority of our approach.
\end{abstract}

%!TEX root = ./main.tex

\vspace{-0.2in}
\section{Introduction}
\label{submission}
\vspace{-0.05in}

Graph clustering aims to group the nodes into several clusters, and routinely finds itself in applications 
%ranging from biochemical structure analysis (...), community detection (...) to recommender system (...).
ranging from biochemical analysis  to  community detection \cite{DBLP:conf/www/JiaZ0W19,liu2023survey}.
With the advance of graph neural networks \cite{kipf2017semisupervised,velickovic2018graph}, deep graph clustering \cite{devvrit2022s3gc,wang2023gc} achieves remarkable success in recent years.

\begin{figure}[t]
\centering
\vspace{-0.1in}
\includegraphics[width=0.97\linewidth]{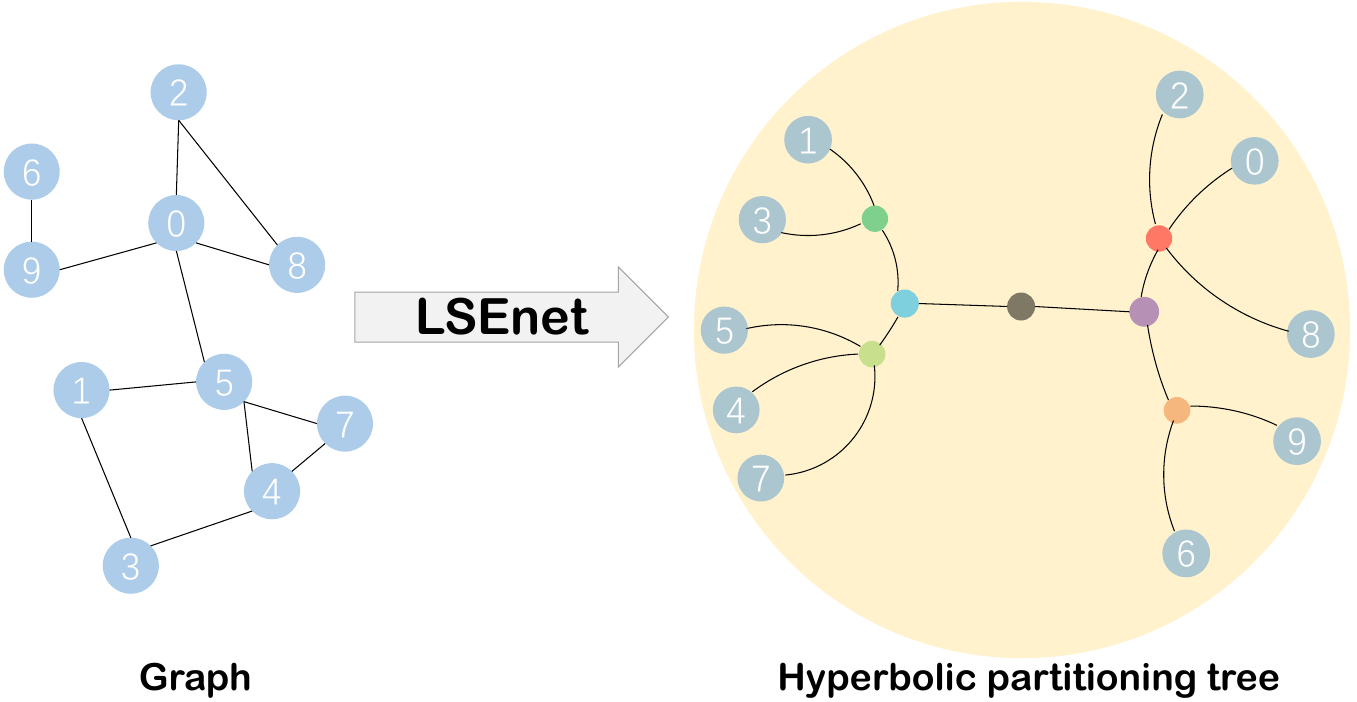}
\vspace{-0.1in}
\caption{Overview. In hyperbolic space, the proposed \texttt{LSEnet} learns a partitioning tree for node clustering without  predefined $K$.}
\label{fig:intro}
\vspace{-0.1in}
\end{figure}

So far, deep graph clustering still cannot work without a predefined cluster number $K$, 
%which is usually unavailable in the real scenario.
and thus one needs to correctly predict the cluster number before clustering, which is often impractical in real cases.
%indeed another challenging problem.
Also, estimating cluster numbers is nontrivial. Empirical methods such as Elbow or Bayesian information criterion need to train the deep model repeatedly \cite{DBLP:journals/sigkdd/Schubert23}, and computation cost is much too expensive.
For the clustering without graph structures, possible solutions free of $K$ include Bayesian non-parametric methods \cite{gershman2011tutorial},  density-based models, e.g., DBSCAN \cite{esterdensitybased}, and hierarchical clustering \cite{cohen-addad2019hierarchical}. 
However, they cannot be directly applied to graphs owing to the inter-correlation among the nodes.
That is, \emph{the problem of graph clustering with unknown cluster number largely remains open.}

In this work, we present a fresh perspective of information theory.
Stemming from Shannon entropy, \textbf{structural information} \cite{li2016structural} is formulated to measure the uncertainty on graphs.
%Accordingly, an essential tree of the graph's self-organization can be decoded via a greedy algorithm, uncovering the cluster structures. 
Minimizing the structural information, an optimal partitioning tree is constructed to describe the graph's self-organization without the knowledge of cluster number.
It sheds light on the targeted problem but presents several significant gaps to deep clustering in the meantime.
First and foremost, the clustering ability of structural entropy is still unclear. 
%To the best of our knowledge
In the literature, structural information has not yet been introduced to deep clustering, though it has been receiving research attention recently   \cite{liu2019rem,wu2023sega,zou2023segslb}. 
Second, the discrete formulation prevents the gradient backpropagation, posing a fundamental challenge to train a deep model.
%Second, the complexity makes the exact computation prohibited. Concretely, the complexity increases exponentially as the dimension increases, and yields $O(N^2\log^2N)$ for sparse graphs in  $3-$dimensional case, where $N$ is the number of nodes.
Third, the classic definition neglects the node features, which are often equally important to graph clustering.

% In light of the issues above, we first establish a new theory of \emph{Differentiable Structure Entropy} (DSE), in order to generalize structural entropy  to the continuous realm....
% Second, motivated by the correspondence between the tree and hyperbolic space, we design a novel Lorentz Structure Entropy neural network (\texttt{LSEnet}) in \emph{hyperbolic space}.  ...
% Main contributions are summarized as follows.

% In light of the issues above, 

% we first establish a novel \emph{Differentiable Structure Entropy} (DSE), to generalize structural entropy to the continuous realm. 

% Second, motivated by the correspondence between the tree and hyperbolic space, we design a novel Lorentz Structure Entropy neural network (\texttt{LSEnet}) in \emph{hyperbolic space}. 

In light of the aforementioned issues, 
%In light of the issues above, 
we present a novel \emph{differentiable structural information} (DSI) in the continuous realm. 
DSI is formulated with level-wise assignment matrices and is equivalent to the classic formulation under binary assignment.
 %(Theorem \ref{theorem.equivalence}).
In fact, the conductance for graph clustering is inherently related to DSI.
The intuition is that in the partitioning tree $\mathcal T$ of DSI minimization, 
the densely connected nodes in the graph tend to be assigned to the same parent node, minimizing the conductance and presenting the cluster structure.
Thus DSI emerges as a new graph clustering objective, not requiring the predefined cluster number.
Next, we consider a partitioning tree $\mathcal T_{\operatorname{net}}$ learned by a neural network, given that our objective supports gradient backpropagation for the network.
We show that the structural entropy of the optimal $\mathcal T_{\operatorname{net}}$ well approximates that of \citet{li2016structural} under slight constraint.
% is well approximated by the tree inferred from our DSI (Theorem \ref{theorem.approx}).
% Minimizing DSI,  
% Intuitively, 
% Theoretically,  we show that DSI is able to serve as a graph clustering objective, not requiring the predefined cluster number (Theorem \label{theorem.conductance}).
%Furthermore, we explore the alignment between the tree and \textbf{hyperbolic space}, 
Furthermore, we design a Lorentz Structural Entropy neural net (\texttt{LSEnet}) to learn $\mathcal T_{\operatorname{net}}$ (as shown in Figure \ref{fig:intro}) in the Lorentz model of \textbf{hyperbolic space}, 
where we further integrate node features to structural information via graph convolution net.
Specifically, we first embed the leaf nodes of the partitioning tree, and then recursively learn the parent nodes from bottom to top (root node), where 
the level-wise parent assignment is attentively determined by the proposed Lorentz assigner.
Consequently, \texttt{LSEnet} combines the advantages of both structural entropy and hyperbolic space for graph clustering.
%, and we derive the close-form expression of parent node embedding.
% With the objective of DSI, \texttt{LSEnet} learns the optimal partitioning tree, discovering the self-organization and clustering structure of the graph.
% We give a new formula for the $H$-dimensional structural information, which not only simplifies the original formula and makes it easy to relax to the continuous version, but also decouples the dependency relationship between tree nodes, enabling us to build a more general theorem about normalized $H$-dimensional structural entropy and reveal its relationship with graph clustering. 
% Moreover, we have proved that DSE is an approximation to the traditional structural entropy, which theoretically guarantees the process is guided correctly.
% Considering tree node embeddings and clustering structure simultaneously, we convert graph nodes with attributes into hyperbolic space and construct a tree in a bottom-up manner. To tackle the unknown internal level node numbers, we build a relaxed partitioning tree in the constructing process. 
% Moreover, we design a decoding algorithm recovering the relaxed tree to the strict one and use a merge-split strategy to deliver the clustering results for given clustering numbers.
% Finally, to evaluate the effectiveness of the proposed methods, we conduct extensive experiments on real-world benchmark datasets for graph clustering and visualize the results of the partitioning tree. Meanwhile, we compare the traditional structural entropy to give intuitive results about approximation and efficiency.
The main contributions are listed as follows.
\begin{itemize}
\item We study a challenging yet practical problem of graph clustering with unknown cluster number and, to our best knowledge, make the first attempt to bridge deep graph clustering and structural information.
\item We present the differentiable structural information (DSI), generalizing the classic theory to the continuous realm.  DSI emerges as a new graph clustering objective, not requiring the cluster number.
\item We design a novel hyperbolic \texttt{LSEnet} with graph neural network, further integrating the structural information and node features. Extensive empirical results show the superiority of \texttt{LSEnet} for graph clustering.
\end{itemize}

\section{Related Work}

\paragraph{Deep Graph Clustering.}
Deep models have achieved state-of-the-art results in node clustering.
Classic methods leverage a reconstructive loss to learn node representations, while identifying node clusters with distance-based or model-based algorithms (e.g., k-means, Gaussian mixture model) \cite{xie2016unsupervised,almahairi2016dynamic,yang2017graph}.
Other methods learn cluster assignment with generative adversarial nets \cite{DBLP:conf/ijcai/0002WGWCG20,DBLP:conf/www/JiaZ0W19}.
Contrastive clustering explores the similarity on the graph itself, pulling positive samples together and pushing negative ones apart \cite{devvrit2022s3gc,pan2021multi,DBLP:conf/www/LiJT22}. 
\citet{sun2023contrastive} consider contrastive graph clustering in the product manifold with a loss of Ricci curvature.
Normalizing flows have recently been introduced to graph clustering \cite{wang2023gc}.
%\citet{sun2023contrastive} introduce the geometric notion of curvature to deep clustering.
 Few deep model is  built without the predefined cluster number, to the best of our knowledge.
 Very recently, \citet{liu2023reinforcement} focus on automatically estimating the number of clusters via reinforcement learning, which is orthogonal to our study.
 %, and obtain clustering results with off-the-shelf algorithms (e.g., k-means)
 
\vspace{-0.1in}

\paragraph{Structural Entropy.}
Information entropy is a key notation of information theory \cite{6773024}, measuring the amount of information for unstructured data, and it fails to study the information on graphs.
On information measure for graphs, early practices, e.g., Von Neumann entropy \cite{bra06} and Gibbs entropy \cite{bia09}, are still defined by unstructured probability distributions, and the graph structure is degenerated.
Recently, structural entropy is proposed in account of the natural self-organizing in the graphs \cite{li2016structural}, and has been successfully applied to graph pooling \cite{wu2022structural}, adversarial attack \cite{liu2019rem}, contrastive learning \cite{wu2023sega}, dimension estimation \cite{yang2023minimum} and graph structural learning \cite{zou2023segslb}. 
% We observe that the methods above are organized in a loose pipeline, i.e., structural entropy is first calculated  independently due to its discrete nature, and then the 
However, structural entropy has not yet been introduced to deep clustering, and the gap roots in the discrete formulation of \citet{li2016structural}.
Besides,  it falls short of considering node features that are important to graph clustering as well.

\vspace{-0.1in}

\paragraph{Riemannian Graph Learning.}
Euclidean space has been the workhorse of graph learning for decades \cite{kdd14DeepWalk,kipf2017semisupervised,velickovic2018graph}. 
In recent years, Riemannian manifolds have emerged as an exciting alternative.
Hyperbolic models \cite{nips17NickelK,chami2019hyperbolica,HVGNN,SunL22CIKM,SunL23AAAI,zhang2021lorentzian,icml18revisitHyperbolic,DBLP:conf/www/FuWSYWPL23} achieve remarkable success on the graphs   dominated by tree-like structures \cite{DBLP:journals/corr/abs-1006-5169}.
In fact, an arbitrary tree can be embedded in hyperbolic space with bounded distortion \cite{sarkar2012low}.
\citet{DBLP:conf/icdm/Fu0WSJWTPY21} study the optimal curvature of hyperbolic space for graph embedding.
Beyond hyperbolic space, the product manifolds \cite{aaai22SelfMix,SunL24AAAI} show its superiority on generic graph structures.
Recently, Ricci curvature of Riemannian geometry is given a differentiable surrogate for graph structural learning \cite{SunL23ICDM}.
Manifold vector fields (ordinary differential equations) \cite{SunL24WWW} are introduced to study information diffusion on the graphs \cite{SunL24SIGIR}.
Note that, \emph{the inherent connection between the tree and hyperbolic space supports the construction of our model.}

\section{Preliminaries and Notations}

Herein, different from the typical setting of existing works, we are interested in a more challenging problem of \emph{graph clustering with unknown cluster number.}
Some preliminary concepts and notations are introduced here.

\paragraph{Graph and Graph Clustering.}
A weighted graph $G$ is represented as $G=(\mathcal{V}, \mathcal{E}, \mathbf X)$. 
$\mathcal{V}$ is the set of $N$ nodes, and the degree of node $v_i$ is denoted as $d_i$.
$\mathcal{E}  \subset \mathcal{V} \times \mathcal{V}$ is the edge set with a weight function $w$, and the edge weights are collected in the adjacency matrix $\mathbf A \in \mathbb R^{N \times N}$.
$\mathbf X$ is the matrix of node features.
 %of binary values. 
For a node subset $\mathcal{U}  \subset \mathcal{V}$,  its volume $\operatorname{Vol}(\mathcal{U})$ is defined as
 the sum of edge weights of the nodes in  $\mathcal{U}$.
Graph clustering aims to group nodes into several clusters.
In the literature, deep clustering methods typically rely on the predefined cluster number $K$ to build the deep model.
In this paper, we consider a more challenging case that \emph{the number of clusters $K$ is unavailable.}
%unknown in line with the practical scenario.

%so that node similarity within the same cluster is greater than that between different clusters.

\paragraph{Hyperbolic Space.} 
%Unlike the flat Euclidean space, hyperbolic space is an isotropic Riemannian manifold which is negatively curved.
%n isotropic Riemannian manifold which is negatively curved.
In Riemannian geometry, unlike the ``flat'' Euclidean space, hyperbolic space is a curved space with negative curvatures.
The notion of curvature $\kappa$ measures the extent of how the manifold deviates from being flat.
We utilize the \emph{Lorentz model}  of hyperbolic space.
Concretely, a $d-$dimensional Lorentz model $\mathbb L^{\kappa, d}$ with curvature $\kappa$
is defined on the manifold of $\{\boldsymbol x \in \mathbb R^{d+1}| \langle \boldsymbol x, \boldsymbol x \rangle_{\mathbb L}= \frac{1}{\kappa}\}$, where the Minkowski inner product $\langle \boldsymbol x, \boldsymbol y \rangle_{\mathbb L}=\boldsymbol x\mathbf R\boldsymbol y $ is defined with the matrix of Riemannian metric $\mathbf R=diag(-1, 1, 1, \cdots, 1) \in \mathbb R^{(d+1)\times(d+1)}$.
$\forall \boldsymbol x, \boldsymbol y$ in the Lorentz, the distance is given as $d_{\mathbb L}(\boldsymbol x, \boldsymbol y)=arccosh(\langle \boldsymbol x, \boldsymbol y \rangle_{\mathbb L})$.
Lorentz norm is defined as $\lVert \boldsymbol{u} \rVert_\mathbb{L} = \sqrt{\langle \boldsymbol u, \boldsymbol u \rangle_\mathbb{L}}$, where $\boldsymbol u$ is a vector in the tangent space.
Please refer to Appendix B for further facts.

% \footnote{All the formulation is given under the standard curvature of $-1$, and hyperbolic spaces with different curvatures are the same in essence \cite{petersen2016riemannian}.} 

Throughout this paper, the lowercase boldfaced $\boldsymbol x$ and uppercase $\mathbf X$ denote vector and matrix, respectively. The notation table is given in Appendix \ref{notationTable}.

\section{Differentiable Structural Information}

% In this section, we first establish a theory of \emph{the improved formula for structural information} to generalize the classic theory \cite{li2016structural}, secondly we give a continuous relaxation to the differentiable realm, yielding a new objective of graph clustering without the cluster number.

In this section, we first establish a new formulation of structural information \cite{li2016structural}, and then give a continuous relaxation to the differentiable realm, yielding a new objective and an optimization approach for graph clustering without the predefined cluster number.
We start with the classic formulation as follows. 
\begin{definition}[$H$-dimensional Structural Entropy \cite{li2016structural}]
\label{hse}
    Given a weighted graph $G=(\mathcal{V}, \mathcal{E})$ with weight function $w$ and a partitioning tree $\mathcal{T}$ of $G$ with height $H$,
     the \emph{structural information} of $G$ with respect to each non-root node $\alpha$ of $\mathcal{T}$ is defined as
    \begin{align}
    \label{eq.si}
        \mathcal{H}^{\mathcal{T}}(G;\alpha)=-\frac{g_\alpha}{\operatorname{Vol}(G)}\log_2\frac{V_\alpha}{V_{\alpha^-}}.
    \end{align}
    In the \emph{partitioning tree} $\mathcal{T}$ with root node $\lambda$, each tree node $\alpha$ is associated with a subset of $\mathcal V$, denoted as module $T_\alpha$, and the immediate predecessor of it is written as $\alpha^-$. The module of the leaf node is a singleton of the graph node. 
% For all the tree nodes in the same level, the union of their modules is equal to the node set of the graph $\mathcal{V}$.
%     where $T_\alpha$ is the subset of $\mathcal{V}$ corresponding to tree node $\alpha$, 
    The scalar $g_\alpha$ is the total weights of graph edges with exactly one endpoint in module $T_\alpha$.
    Then, the $H$-dimensional structural information of $G$ by $\mathcal{T}$ is given as,
    \begin{align}
    \label{si}
        \mathcal{H}^{\mathcal{T}}(G) = \sum\nolimits_{\alpha \in \mathcal{T}, \alpha \neq \lambda }\mathcal{H}^{\mathcal{T}}(G;\alpha).
    \end{align}
    Traversing all possible partitioning trees of $G$ with height $H$, \emph{$H$-dimensional structural entropy} of $G$ is defined as
    \begin{align}
        \mathcal{H}^{H}(G) = \min_\mathcal{T} \mathcal{H}^{\mathcal{T}}(G), \quad
    \label{eq.opt_tree}
        \mathcal{T}^*=\arg_{\mathcal{T}} \min \mathcal{H}^{\mathcal{T}}(G),
    \end{align}
    where  $\mathcal{T}^*$ is the optimal tree of $G$ which encodes the self-organization and minimizes the uncertainty of the graph.
\end{definition}
% Specially, as $H=1$, the one-dimensional structural entropy of $G$
% \begin{align}
%   \mathcal{H}^1(G)=-\frac{1}{\operatorname{Vol}(G)}\sum_{i=1}^N d_i\log_2\frac{d_i}{\operatorname{Vol}(G)}  
% \end{align}
% is called the positioning entropy of graph $G$.

The structural information in Eq. (\ref{si}) is formulated node-wisely, and cannot be optimized via gradient-based methods.

% There are three major shortcomings: 1) the discrete formulation prevents the gradient backpropagation for deep learning, 2) the complexity makes the exact computation prohibited,  and 3) the classic definition neglects the node features, which are equally important to the graph structures.

\subsection{A New Formulation}

To bridge this gap, we present a new formulation of structural information with the level-wise assignment, which is shown to be equivalent to the classic formulation in Eq. (\ref{si}).

\begin{definition}[Level-wise Assignment]
\label{def.C}
    %Consider a graph $G$ with $N$ nodes, and 
    For a partitioning tree $\mathcal{T}$ with height $H$, assuming that the number of tree nodes at the $h$-th level  is $N_h$, we define a \emph{level-wise parent assignment matrix $\mathbf{C}^h \in \{0, 1\}^{N_h \times N_{h-1}}$ from $h$-th to $(h-1)$-th level}, where $\mathbf{C}^h_{ij}=1$ means the $i$-th node of $\mathcal{T}$ at $h$-th level is the parent node of $j$-th node at $(h-1)$-th level.
\end{definition}
\begin{definition}[H-dimensional Structural Information]
\label{DSE}
    For a graph $G$ and its partitioning tree $\mathcal{T}$ in Definition \ref{def.C}, we rewrite the formula of $H$-dimensional structural information of $G$ with respect to $\mathcal{T}$ \textbf{at height $h$} as 
    \begin{equation}
        \mathcal{H}^\mathcal{T}(G;h)=-\frac{1}{V} 
            \sum\limits_{k=1}^{N_h}(V^h_k - \sum\limits_{(i,j)\in \mathcal{E}}S^h_{ik}S^h_{jk}w_{ij})\log_2\frac{V^h_k}{V^{h-1}_{k^-}}
    \label{kse}
    \end{equation}
    where $V=\operatorname{Vol}(G)$ is the volume of $G$. For the $k$-th node in height $h$, $V^h_k$ and $V^{h-1}_{k^-}$ are the volume of graph node sets $T_k$ and $T_{k^-}$, respectively. Thus we have
    \begin{align}
    \label{eq.s}
        \mathbf{S}^h &= \prod\nolimits_{k=H+1}^{h+1}\mathbf{C}^k,  \quad \mathbf{C}^{H+1}=\mathbf{I}_N,\\
        V^h_k &= \sum\nolimits_{i=1}^N S^h_{ik}d_i, \quad V^{h-1}_{k^-} = \sum\nolimits_{k'=1}^{N_{h-1}}\mathbf{C}_{kk'}^h V^{h-1}_{k'} .
         \label{eq.cv}
    \end{align}
    Then, the $H$-dimensional structural information of $G$ is
       $ \mathcal{H}^\mathcal{T}(G) = \sum\nolimits_{h=1}^H H^\mathcal{T}(G;h)$.
\end{definition}
\begin{theorem}[Equivalence]
   \label{theorem.equivalence}
    The formula $\mathcal{H}^\mathcal{T}(G)$ in Definition \ref{DSE} is equivalent to Eq. (\ref{si}) given in Definition \ref{hse}.
\end{theorem}
\begin{proof}
Please refer to Appendix \ref{proof.equivalence}.
\end{proof}
% \begin{align}
%     \label{eq.opt_z}
%     \mathbf{Z}^* &= \mathop{\arg\min}_{Z} \mathcal{H}_{hyp}^H(G; \mathbf{Z})  \\
%     \label{eq.opt_htree}
%     \mathcal{T}_\mathbb{H}^* &= \operatorname{decode}(\mathbf{Z}^*)
% \end{align}

\subsection{Properties}

%In this part, we first give the bound between our DSE and the classic formulation, then show the connection between DSE and graph clustering.
This part shows some general properties of the new formulation and theoretically demonstrates the inherent connection between structural entropy and graph clustering. 
We first give an arithmetic property regarding Definition \ref{DSE} to support the following claim on graph clustering.
The proofs of the lemma/theorems are detailed in Appendix A.
\begin{lemma}[Additivity-Appendix \ref{proof.equivalence}]
\label{lemma.add}
    %Given a partitioning tree $\mathcal{T}$ with height $H$, 
    The $1$-dimensional structural entropy of $G$ can be decomposed as follows
    \begin{align}
        \mathcal{H}^1(G) = \sum_{h=1}^H\sum_{j=1}^{N_{h-1}}\frac{V^{h-1}_j}{V}E([\frac{C^h_{kj}V^h_k}{V^{h-1}_j}]_{k=1,...,N_h}),
    \end{align}
    where 
      $  E(p_1, ..., p_n) = -\sum_{i=1}^n p_i\log_2 p_i$
 is the entropy. 
\end{lemma}
\begin{theorem}[Connection to Graph Clustering-Appendix \ref{proof.conductance}]
\label{theorem.conductance}
        Given a graph $G=(\mathcal{V}, \mathcal{E})$ with  $w$, the normalized $H$-structural entropy of graph $G$ is defined as
    $\tau(G;H) = {\mathcal{H}^H(G)}/{\mathcal{H}^1(G)}$, and 
$\Phi(G)$ is the graph conductance. 
    With the additivity \emph{(Lemma \ref{lemma.add})}, the following inequality holds,
    %Then the normalized $H$-dimensional structural information of $G$ w.r.t. a partitioning tree $\mathcal{T}$ satisfies the following property:
    \begin{align}
        \tau(G; H) \geq \Phi(G).
    \end{align}
\end{theorem}
\begin{proof}
    We show the key equations here, and further details are given in Appendix \ref{proof.conductance}. Without loss of generality, we assume $\min\{ V^h_k, V-V^h_k \}=V^h_k$. From Definition \ref{DSE},
    \begin{align}
        \mathcal{H}^\mathcal{T}(G) &= -\frac{1}{V}\sum_{h=1}^H\sum_{k=1}^{N_h} \phi_{h,k}V^h_k\log_2\frac{V^h_k}{V^{h-1}_{k^-}} \nonumber \\
        &\geq -\frac{\Phi(G)}{V}\sum_{h=1}^H \sum_{k=1}^{N_h}V^h_k\log_2\frac{V^h_k}{V^{h-1}_{k^-}},
    \end{align}
    Let the $k$-th tree node in height $h$ denoted as $\alpha$, 
    then $\phi_{h,k}$ is the conductance of graph node subset $T_{\alpha}$, 
    and is defined as
    $\frac{\sum_{i\in T_{\alpha}, j\in \bar{T}_{\alpha}}w_{ij}}{\min\{ \operatorname{Vol}(T_{\alpha}), \operatorname{Vol}(\bar{T}_{\alpha}) \}}$.
    where $\bar{T}_{\alpha}$ is the complement set of $T_{\alpha}$. 
    Thus, graph conductance is  given as $\Phi(G) = \min_{h,k}\{\phi_{h,k}\}$. 
    Next, we utilize Eq. (\ref{eq.cv}) to obtain the following inequality
    \begin{align}
        &\mathcal{H}^\mathcal{T}(G) \geq -\frac{\Phi(G)}{V}\sum_{h=1}^H \sum_{k=1}^{N_h}\sum_{j=1}^{N_{h-1}}C^h_{kj}V^h_k\log_2\frac{V^h_k}{V^{h-1}_{k^-}} \nonumber \\
        &=-\frac{\Phi(G)}{V}\sum_{h=1}^H \sum_{j=1}^{N_{h-1}} V^{h-1}_j \sum_{k=1}^{N_h} \frac{C^h_{kj}V^h_k}{V^{h-1}_j}\log_2\frac{C^h_{kj}V^h_k}{V^{h-1}_j} \nonumber \\
        &=\Phi(G)\sum_{h=1}^H\sum_{j=1}^{N_{h-1}}\frac{V^{h-1}_j}{V}E([\frac{C^h_{kj}V^h_k}{V^{h-1}_j}]_{k=1,...,N_h}) \nonumber \\
        &=\Phi(G)\mathcal{H}^1(G).
    \end{align}
Since $\tau(G;\mathcal{T}) = \frac{\mathcal{H}^\mathcal{T}(G)}{\mathcal{H}^1(G)} \geq \Phi(G)$ holds for every $H$-height partitioning tree $\mathcal{T}$ of $G$, $\tau(G)\geq \Phi(G)$ holds.
\end{proof}
The conductance is a well-defined objective for graph clustering \cite{DBLP:conf/kdd/YinBLG17}.
%The conductance of graph $\Phi(G)$ is given as $\min\nolimits_{\forall P \subset \mathcal{V}} \Phi(P)$.
%     Given a weighted graph $G=(\mathcal{V}, \mathcal{E})$ with a weight function $w$, for a subset $P \subseteq \mathcal{V}$, the conductance of $P$ is
% \begin{align}
%     \Phi(P) = \frac{\sum_{i\in P, j\in \bar{P}}w_{ij}}{\min\{ \operatorname{vol}(P), \operatorname{Vol}(\bar{P}) \}}
% \end{align}
%That is, , given $1-$dimensional structural entropy is a constant. 
$\forall G$, the one-dimensional structural entropy
  $\mathcal{H}^1(G)=-\frac{1}{\operatorname{Vol}(G)}\sum_{i=1}^N d_i\log_2\frac{d_i}{\operatorname{Vol}(G)} $ 
is yielded as a constant.
% the positioning entropy of graph $G$.{}
Minimizing Eq. (\ref{kse}) is thus equivalent to minimizing conductance, grouping nodes into clusters. 
%Besides the clustering ability, our forDSE is formulated without the knowledge of cluster number (Definition \ref{DSE}).
Also, Eq. (\ref{kse}) needs no knowledge of cluster number in the graph $G$.
\emph{In short, our formulation is capable of serving as an objective of graph clustering without predefined cluster numbers.}

\subsection{Differentiablity \& Deep Graph Clustering}

% the optimization of structural information for graph clustering.

% The gradient over  $\mathbf C$ exists in Eq. (\ref{kse}) , and thus our  formulation supports gradient backpropagation with respect to the elements of assignment matrix.
% %of structural information is differentiable if those elements are differentiable.
% It is interesting to consider assignment element is parameterized by a neural net.

% In this section, we give a framework for optimizing structural information in a differentiable way by our proposed model \texttt{LSEnet}, a hyperbolic neural model to uncover the self-organizing pattern of a graph. The detailed introduction of \texttt{LSEnet} is described in the next section.

% From Definition \ref{DSE} and Theorem \ref{theorem.equivalence}, we can find that if all the elements of matrix $C^h$ are differentiable, the $H$-dimensional structural information is also differentiable. Following this discovery, we can deliver the $C^h_{ij}$ by our proposed \texttt{LSEnet} which can be optimized by minimizing structural information w.r.t. an unknown tree.

Here, we elaborate on how to use a deep model to conduct graph clustering with our new objective.
Recall Eq. (\ref{kse}), our objective is differentiable over the parent assignment.
If the assignment is relaxed to be the likelihood given by a neural net,
our formulation supports gradient backpropagation to learn the deep model.
For a graph $G$, we denote the partitioning tree as $\mathcal T_{\operatorname{net}}$ where the mapping from learnable embeddings $\mathbf{Z}$ to the assignment is done via a neural net.
The differentiable structural information (DSI) is given as $\mathcal{H}^{\mathcal T_{\operatorname{net}}}(G; \mathbf{Z}; \mathbf \Theta)$. It takes the same form of Eq. (\ref{kse}) where the assignment is derived by a neural net with parameters $\Theta$.
Given $G$, we consider DSI minimization as follows, 
 \begin{align}
 \label{eq.opt_z}
    \mathbf{Z}^*, \mathbf \Theta^* = \arg_{\mathbf{Z},\mathbf \Theta} \min \mathcal{H}^{\mathcal T_{\operatorname{net}}}(G; \mathbf{Z}; \mathbf \Theta).
 \end{align}
 %where $\Theta$ is the parameters of $\texttt{LSEnet}$.
% We can decode the corresponding partitioning tree from the embeddings as
%  \begin{align}
%  \label{eq.opt_tree}
%      \mathcal{T}_\mathbb{L}^* = \operatorname{decode}(\mathbf{Z}^*).
%      \vspace{-0.1in}
%  \end{align}
% denotes as , 
% Then, the optimal embeddings for tree nodes are given by
%Minimizing Eq. (\ref{eq.opt_z}), 
%Consequently, we derive the level-wise assignment with $\Theta^*$, and thus

We learn node embeddings $\mathbf{Z}^*$ and parameter $\mathbf \Theta^*$ to derive the level-wise parent assignment for $G$.
Then, the optimal partitioning tree $\mathcal{T}_{\operatorname{net}}^*$ of $G$ is constructed by the assignment,
so that \emph{densely connected nodes have a higher probability to be assigned to the same parent in $\mathcal{T}_{\operatorname{net}}^*$, minimizing the conductance and presenting the cluster structure.}
Meanwhile, node embeddings jointly learned in the optimization provide geometric notions to refine clusters in representation space.

Next, we introduce the theoretical guarantees of $\mathcal{T}_{\operatorname{net}}^*$. 
We first define the notion of equivalence relationship in partitioning tree $\mathcal{T}$ with height $H$. 
If nodes $i$ and $j$ are in the same module at the level $h$ of $\mathcal{T}$ for $h=1, \cdots, H$, 
    they are said to be in equivalence relationship, denoted as $i \overset{h}{\sim} j$.

\begin{theorem}[Bound-Appendix \ref{proof.theoremapprox}]
\label{theorem.approx}
 %For any graph $G$, $\mathcal{T}^*$ is the optimal partitioning tree given by  \citet{li2016structural},
% and $\mathcal{T}_{\operatorname{net}}^*$ with height $H$ is the partitioning tree given from Eq. (\ref{eq.opt_z}).
  For any graph $G$, $\mathcal{T}^*$ is the optimal partitioning tree  of  \citet{li2016structural},
 and $\mathcal{T}_{\operatorname{net}}^*$ with height $H$ is the partitioning tree given by Eq. (\ref{eq.opt_z}).
For any pair of leaf embeddings $\boldsymbol z_i$ and $\boldsymbol z_j$ of $\mathcal{T}_{\operatorname{net}}^*$, 
 there exists bounded real functions $\{f_h\}$ and constant $c$,
 such that for any $0 < \epsilon < 1$, $\tau \leq \mathcal{O}(1/ \ln[(1-\epsilon)/\epsilon])$ and  $\frac{1}{1 + \exp\{-(f_h(\boldsymbol z_i, \boldsymbol z_j) - c) / \tau\}} \geq 1 - \epsilon$ satisfying $i\overset{h}{\sim} j$, we have 
    \begin{align}
        \lvert \mathcal{H}^{\mathcal{T}^*}(G) - \mathcal{H}^{\mathcal{T}_{\operatorname{net}}^*}(G) \rvert \leq \mathcal{O}(\epsilon).
    \end{align}
\end{theorem}
% That is, the difference between structural entropy calculated by our neural approach and \citet{li2016structural} is quite small, and bounded.
% Thus, the optimal partitioning tree $\mathcal{T}^*$ is well approximated by the tree $\mathcal{T}_{\operatorname{net}}^*$ from Eq. (\ref{eq.opt_z}).
That is, the structural entropy calculated by our $\mathcal{T}_{\operatorname{net}}^*$ is well approximated to that of \citet{li2016structural}, and the difference is bounded under slight constraints.
Thus, $\mathcal{T}_{\operatorname{net}}^*$ serves as an alternative to the optimal $\mathcal{T}^*$ for graph clustering.
\begin{theorem}[Flexiblity-Appendix \ref{proof.flex}]
\label{theorem.flex}
    $\forall G$, given a partitioning tree $\mathcal{T}$ and adding a node $\beta$ to get a relaxed $\mathcal{T}'$ , the structural information remains unchanged, $H^{\mathcal{T}}(G)=H^{\mathcal{T}'}(G)$, if  one of the following conditions holds:
    \begin{enumerate}
    %  \item $\beta$ as a leaf node, and the corresponding node subset $T_\beta$ is an empty set.
        \item $\beta$ as a leaf node, and its module $T_\beta$ is an empty set.
      \item $\beta$ is inserted between node $\alpha$ and its children nodes so that the modules  $T_\beta$ and $T_\alpha$ are equal.
\end{enumerate}
\end{theorem}
The theorem above will guide the architecture design of the neural net for $\mathcal{T}_{\operatorname{net}}^*$.

\section{\texttt{LSEnet}}

We propose a novel Lorentz Structural Entropy neural Net (\texttt{LSEnet}), which aims to learn the optimal partitioning tree  $\mathcal{T}_{\operatorname{net}}^*$ in the Lorentz model of \textbf{hyperbolic space}, where we further incorporate node features with structural information by graph convolution net. 
% graph clustering.
%In a nutshell, \texttt{LSEnet} equips graph neural network with our DSE in the Lorentz model of hyperbolic space, further incorporating node features with structural entropy. % graph clustering.
%In hyperbolic space, we build the coding tree in a bottom-up manner, and seamlessly integrate the structural entropy and node features via a Lorentz graph convolution. 
First, we show the reason we opt for hyperbolic space, rather than Euclidean space.
\begin{theorem}[Tree $\mathcal T$  and Hyperbolic Space \cite{sarkar2012low}]
\label{theorem.tree}
% Any tree whose edges length bounded below a constant about $\frac{1+\epsilon}{\epsilon}$ for arbitrary $\epsilon$, can be embedded into hyperbolic space with distortion bounded by $1+\epsilon$.
% For any 
$\forall \mathcal T$, scaling all edges by a constant so that the edge length is bounded below $\nu\frac{1+\epsilon}{\epsilon}$, there exists an embedding in hyperbolic space that the distortion\footnote{The distortion is defined as $\frac{1}{|\mathcal V|^2} \sum_{ij}\left| \frac{d_G(v_i, v_j)}{d(\mathbf x_i, \mathbf x_j)}-1\right|$, where each node $v_i \in \mathcal V$ is embedded as $\mathbf x_i$ in representation space. $d_G$ and $d$ denote the distance in the graph and the space, respectively.} 
overall node pairs are bounded by $1+\epsilon$. \emph{($\nu$ is a constant detailed in Appendix B.)}
\end{theorem}

Hyperbolic space is well suited to embed the partitioning tree, and Theorem \ref{theorem.tree} does not hold for Euclidean space.

\begin{figure}[t]
\centering
\includegraphics[width=1\linewidth]{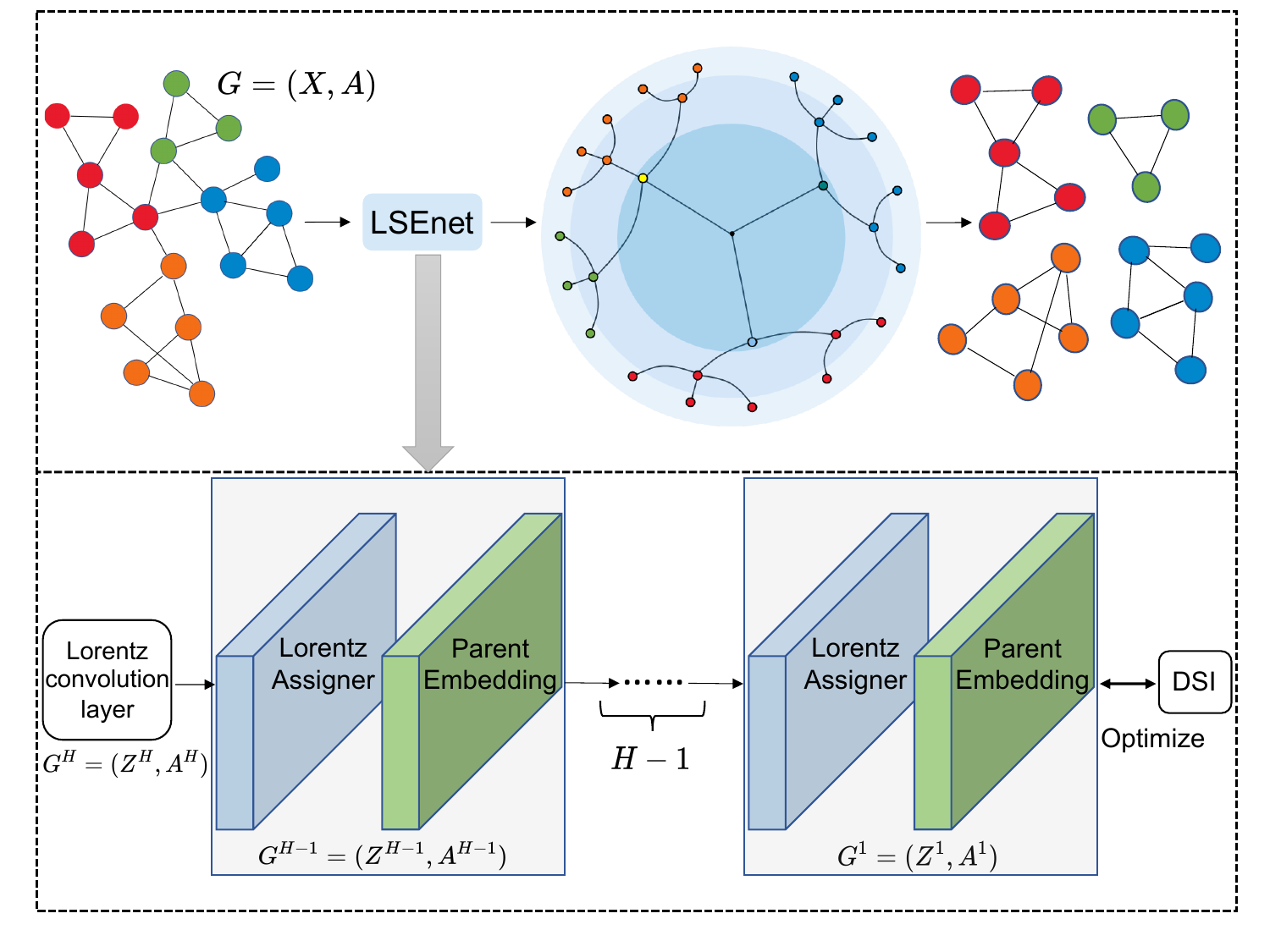}
\vspace{-0.1in}
\caption{The overall architecture of \texttt{LSEnet}. We encode $G$ into Lorentz model $\mathbb{L}^{\kappa, d_{\mathcal T}}$ via a Lorentz convolution layer, and recursively utilize Lorentz assignor and geometric centroid to construct a tree from down to up. Optimizing  DSI, we learn the optimal $\mathcal{T}_{\operatorname{net}}^*$  in hyperbolic space, and obtain clustering results correspondingly. }
\label{fig:lsenet}
\end{figure}

\textbf{Overall architecture} of \texttt{LSEnet} is sketched in Figure \ref{fig:lsenet}.
%In hyperbolic space, we learn the partitioning tree in the bottom-up manner. \texttt{LSEnet} is built up by stacking Lorentz assigners, where each assigner is responsible for a tree layer, \emph{self-supervised by the objective of our DSE (Eq. \ref{kse})}.
In hyperbolic space,  \texttt{LSEnet} first embeds leaf nodes of the tree, and then recursively learns parent nodes, \emph{self-supervised by our new clustering objective Eq. (\ref{kse})}.

%%%%%%%%%%%%%%%%%%%%%%%%%%%%%%%%%%%%%%%%%%%%%%%%%%%%%%%%%%%
%Huang
% From Proposition \ref{DSE} we notice that the $H$-dimensional structural entropy of graph $G$ can be calculated layer by layer from down to top, which means that we can construct the optimal partitioning tree of $G$ layer-wisely in a down-top way.

% The First thing is to get leaf node embeddings, not only need we consider the graph topology and graph node attributes, but also the tree embedding in hyperbolic space. Therefore, we should adopt some transforms in hyperbolic space meanwhile handling them with graph topology.

\subsection{Embedding Leaf Nodes}
\label{leaf}
%Here, we study the 
We design a Lorentz convolution layer to learn leaf embeddings in hyperbolic space, 
$\operatorname{LConv}: \boldsymbol x_i  \to  \boldsymbol z^H_i, \forall v_i$, 
where $\boldsymbol x_i\in \mathbb L^{\kappa, d_0}$  and $\boldsymbol z^H_i \in \mathbb L^{\kappa, d_{\mathcal T}}$ are the node feature and leaf embedding, respectively.
In the partitioning tree of height $H$, 
the level of nodes is denoted by superscript of $h$
%the nodes of $h$-th level node embedding $\boldsymbol z^h_{\cdot}$ is denoted as  $d_h$, 
and we have $h=H, \cdots, 1, 0$ from leaf nodes (bottom) to root (top), correspondingly.

We adopt attentional aggregation in $\operatorname{LConv}$ and specify the operations as follows.
First, dimension transform is done via a \emph{Lorentz linear operator} \cite{chen2022fullyb}.
For node $\boldsymbol x_i\in \mathbb L^{\kappa, d_0}$, the linear operator is given as 
\begin{align}
    \operatorname{LLinear}(\boldsymbol{x}) = 
    \begin{bmatrix}
        \sqrt{\lVert h(\mathbf{W}\boldsymbol{x},\boldsymbol{v}) \rVert^2 - \frac{1}{\kappa}} \ \\
        h(\mathbf{W}\boldsymbol{x},\boldsymbol{v}) \
    \end{bmatrix}
    \in \mathbb L^{\kappa, d_{\mathcal T}},
\end{align}
where $h$ is a neural network, and $\mathbf{W}$ and $\boldsymbol{v}$ are parameters.
Second, we derive attentional weights from the \emph{self-attention mechanism}. 
% With $\mathbf{Q}, \mathbf{K}, \mathbf{V}\in \mathbb L^{\kappa, d_{\mathcal T}}$ given from $\mathbf{X} \in \mathbb L^{\kappa, d_0}$ via $\operatorname{LLinear}$, the attentional weight is calculated as 
% \begin{align}
%     \Omega_{ij} = \operatorname{LAtt}(\mathbf{Q}, \mathbf{K})=
%      \frac{\exp(-  \gamma   d^2_{\mathbb{L}}(\boldsymbol{q}_i, \boldsymbol{k}_j)  )}
%      {\sum_{l=1}^{N}\exp( -  \gamma   d^2_{\mathbb{L}}(\boldsymbol{q}_i, \boldsymbol{k}_l)  )},
%      \label{LAtt}
%          \vspace{-0.07in}
% \end{align}
% where $\gamma$ is $1/\sqrt{N}$, and $\mathbf{X} \in \mathbb L^{\kappa, d}$ denotes a matrix whose row vectors represent the points in hyperbolic space $\mathbb L^{\kappa, d}$.
Concretely, the attentional weight $\omega_{ij}$ between nodes $i$ and $j$ is calculated as 
\begin{align}
    \omega_{ij} = \operatorname{LAtt}(\mathbf{Q}, \mathbf{K})=
     \frac{\exp(-  \frac{1}{\sqrt{N}}   d^2_{\mathbb{L}}(\boldsymbol{q}_i, \boldsymbol{k}_j)  )}
     {\sum_{l=1}^{N}\exp( -   \frac{1}{\sqrt{N}}   d^2_{\mathbb{L}}(\boldsymbol{q}_i, \boldsymbol{k}_l)  )},
     \label{LAtt}
\end{align}
where $\boldsymbol{q}$ and $\boldsymbol{k}$ are the query and key vector collected in row vector of $\mathbf Q$ and $\mathbf K$, respectively. The queries and keys are derived from node feature $\boldsymbol x$ via $\operatorname{LLinear}$ with different parameters.
Third, weighted aggregation is considered as the \emph{arithmetic mean} among manifold-valued vectors.
Given a set of points $\{\boldsymbol{x}_i\}_{i=1, \cdots, N}$ in the Lorentz model $\mathbb L^{\kappa, d_{\mathcal T}}$, 
\begin{align}
    \operatorname{LAgg}(\boldsymbol{\omega}, \mathbf{X}) = \frac{1}{\sqrt{-\kappa}} \sum_{i=1}^N \frac{ \omega_i}{\lvert \lVert \sum_{j=1}^n \omega_j\boldsymbol{x}_j \rVert_\mathbb{L} \rvert} \boldsymbol{x}_i,
     \label{LAgg}
\end{align}
where $\boldsymbol{\omega}$ is the weight vector, and $\boldsymbol x$ are summarized in $\mathbf{X}$.
The augmented form is that for $\mathbf{\Omega}=[\boldsymbol{\omega}_1, \cdots, \boldsymbol{\omega}_N]^\top$, 
we have $\operatorname{LAgg}(\mathbf{\Omega}, \mathbf{X}) = [\boldsymbol \mu_1, \cdots, \boldsymbol  \mu_N]^\top$, 
where the weighted mean is $\boldsymbol  \mu_i =\operatorname{LAgg}(\boldsymbol{\omega}_i, \mathbf{X})$.
Overall, the Lorentz convolution layer is formulated as,
\begin{align}
\operatorname{LConv}(\mathbf{X}|\mathbf{A})=\operatorname{LAgg}(\operatorname{LAtt}(\mathbf{Q}, \mathbf{K})\odot \mathbf{A}, \operatorname{LLinear}(\mathbf{X}) ),
\end{align}
where $\odot$ is the Hadamard product, masking the attentional weight if the corresponding edge does not exist in the graph.

\subsection{Learning Parent Nodes}
\label{parent}

A primary challenge is that, in the partitioning tree, \emph{the node number at each internal level is unknown.}
To address this issue, we introduce a simple yet effective method, setting a large enough node number $N_h$ at the $h$-th level.
A large $N_h$ may introduce redundant nodes and result in a relaxed partitioning tree. According to \textbf{Theorem \ref{theorem.flex}} established in Sec. 4.3,  redundant nodes in the partitioning tree do not affect the value of structural entropy, and finally present as empty leaf nodes by optimizing our objective.
Theoretically, if an internal level has insufficient nodes, the self-organization of the graph can still be described by multiple levels in the partitioning tree. 

%The remaining tasks are to study the assignment between the levels and the embedding of upward level.
Without loss of generality, we consider the assignment between $h$-th and $(h-1)$-th levels given node embeddings $\boldsymbol z^{h}  \in \mathbb L^{\kappa, d_{\mathcal T}}$  at $h$-th level.
%, for $h\in[H, 1]$.
Recalling Definition \ref{def.C}, the $i$-th row of assignment $\mathbf C^h \in \mathbb R^{N_h \times N_{h-1}}$ describes the belonging of $i$-th node at $h$-th to the parent nodes at $(h-1)$-th.
We design a Lorentz assigner following the intuition that neighborhood nodes in the graph tend to have similar assignments.
%Intuitively, neighborhood nodes in the graph tend to have similar assignment.
%We first leverage a Multilayer Perceptron $\operatorname{MLP}$ to learn the assignment from embeddings, 
Concretely, we leverage Multilayer Perceptron $\operatorname{MLP}$ to learn the assignment from embeddings, 
%and meanwhile we utilize  $\operatorname{LAtt}$ defined in Eq. \ref{LAtt} to parameterize the similarity. 
and meanwhile the similarity is parameterized by $\operatorname{LAtt}$ defined in Eq. \ref{LAtt}. 
Thus,  the Lorentz assigner is formulated as
\begin{align}
   \mathbf{C}^h =\sigma\left(  \left( \operatorname{LAtt}(\mathbf{Q}, \mathbf{K}) \odot \mathbf{A}^h\right)\operatorname{MLP}(\mathbf Z^h) \right),
    \label{assignment}
\end{align}
where $\mathbf{A}^h$ is the graph structure at $h$-th level of the tree.
$\sigma$ denotes the row normalization via $Softmax$ function.

The remaining task is to infer the node embeddings $\boldsymbol z^{h-1} \in \mathbb L^{\kappa, d_{\mathcal T}}$ at $(h-1)$-th level. 
As reported in \citet{chami2020trees}, a parent node locates at the point that has the shortest path to all the child nodes at $h$-th level 
and, correspondingly, the parent node is the  Fr\'{e}chet mean of child nodes in the manifold.
The challenge here is that Fr\'{e}chet mean regarding  the canonical distance exists no closed-form solution \cite{DBLP:conf/icml/LouKJBLS20}.
Alternatively, we consider the geometric centroid with respect to the squared Lorentz distance, where the weights are given by the soft assignment $\mathbf{C}^h$,
\begin{align}
    \boldsymbol z^{h-1}_j=\arg_{   \boldsymbol z^{h-1}_j} \min \sum\nolimits_{i=1}^N c_{ij}d^2_{\mathbb L}(\boldsymbol z^{h-1}_j, \boldsymbol z^h_i).
    \label{centroid}
\end{align}
%As a result, the node embeddings are summarized in 
%It is a constrained optimization whose solution is required to be in the manifold $\mathbb L^{\kappa, d_{\mathcal T}}$.
Solving the optimization constrained in the manifold $\mathbb L^{\kappa, d_{\mathcal T}}$, we derive the closed-form solution of parent node embeddings as follows,
\begin{align}
    \mathbf{Z}^{h-1}=\operatorname{LAgg}(\mathbf{C}^h, \mathbf{Z}^h) \in \mathbb L^{\kappa, d_{\mathcal T}},
    \label{parentNode}
\end{align}
which is the augmented form of Eq. (\ref{LAgg}).
\begin{theorem}[Geometric Centroid-Appendix \ref{hyperbolic.midpoint}]
\label{theorem.centroid}
In hyperbolic space $\mathbb L^{\kappa, d_{\mathcal T}}$,
for any set of points $\{\boldsymbol z^{h}_i\}$,
the arithmetic mean of
$\boldsymbol z^{h-1}_j =
\frac{1}{\sqrt{-\kappa}} \sum_{i=1}^N \frac{ c_{ji}}{\lvert \lVert \sum_{l=1}^n c_{jl} \boldsymbol z^{h}_l  \rVert_\mathbb{L} \rvert} \boldsymbol z^{h}_i $
is the manifold $\boldsymbol z^{h-1}_j \in \mathbb L^{\kappa, d_{\mathcal T}}$,
and is the closed-form solution of the geometric centroid defined in the minimization of Eq. \ref{centroid}.
\end{theorem}
% \begin{proof}
% Please refer to Appendix B.4.
% \vspace{-0.1in}
% \end{proof}
\paragraph{Remark.}
In fact, the geometric centroid in Theorem \ref{theorem.centroid} is also equivalent to the gyro-midpoint in Poincar\'{e} ball model of hyperbolic space, detailed in Appendix \ref{hyperbolic.midpoint}

%Meanwhile, the graph structure among the parent nodes at $(h-1)$-th level in the partitioning tree is given as follows,
According to Eq. (\ref{assignment}), it requires the graph structure  $\mathbf{A}^{h-1}$ among the parent nodes at $(h-1)$-th level to derive the assignment. We give the adjacency matrix of the $(h-1)$-th level graph as follows,
\begin{align}
    \mathbf{A}^{h-1} &= (\mathbf{C}^h)^\top \mathbf{A}^h \mathbf{C}^h.
    \label{structure}
\end{align}
We recursively utilize Eqs. (\ref{assignment}), (\ref{parentNode}) and (\ref{structure}) to build the partitioning tree from bottom to top in hyperbolic space.

\begin{algorithm}[t]
    \caption{Training \texttt{LSEnet}}
    \label{alg:lsenet}
    \renewcommand{\algorithmicrequire}{\textbf{Input:}}
    \renewcommand{\algorithmicensure}{\textbf{Output:}}
    \begin{algorithmic}[1]
        \REQUIRE A weighted graph $G=(\mathcal{V}, \mathcal{E}, \mathbf{X})$, Height of partitioning tree $H$, Training iterations $L$
         \ENSURE The partitioning tree $\mathcal{T}$; Tree nodes embeddings $\{\mathbf{Z}^h\}_{h=1,...,H}$ at all the levels;
        \FOR{$epoch=1$ to $L$}
            \STATE Obtain leaf node embeddings $\mathbf{Z}^H = \operatorname{LConv}(\mathbf{X}, \mathbf{A})$.
            \FOR{$h=H-1$ to $1$}
                \STATE Compute $\mathbf{Z}^h$, $\mathbf{A}^h$, $\mathbf{C}^h$ with Eqs. \ref{assignment}, \ref{parentNode} and \ref{structure}.
               \STATE Compute $\mathbf{S}^h$ for $h=1, \cdots, H$ in Eq. \ref{eq.s}.
            \ENDFOR
        \STATE Compute the objective of DSI in Eq. \ref{kse}.
        \STATE Optimize parameters via Riemannian Adam.
        \ENDFOR
        \STATE Create a root node $\lambda$ and a queue $\mathcal{Q}$.
        \WHILE{$\mathcal{Q}$ is not empty}
            \STATE Get first item $\alpha$ in $\mathcal{Q}$.
            \STATE Let $h=\alpha.h+1$ and search subsets $P$ from $\mathbf{S}^h$.
            \STATE Create nodes from $P$ and put into the queue $\mathcal{Q}$.
            \STATE Add theses nodes into $\alpha$'s children list.
        \ENDWHILE
        \STATE Return the partitioning tree $\mathcal{T}:=\lambda$.
    \end{algorithmic}
\end{algorithm}

%\hfill $\rhd$ \emph{Training the Discriminator}\\

% \begin{algorithm}[t]
%            \caption{Training \texttt{LSEnet}}
%     \label{alg:lsenet}
%         \KwIn{A weighted graph $G=(\mathcal{V}, \mathcal{E}, \mathbf{X})$, Height of partitioning tree $H$, Training iterations $L$}
%         \KwOut{The partitioning tree $\mathcal{T}$; Tree nodes embeddings at all the levels $\{\mathbf{Z}^h\}_{h=1,...,H}$}
%             \For{$epoch=1$ to $L$} {
%                 Obtain leaf node embeddings $\mathbf{Z}^H = \operatorname{LConv}(\mathbf{X}, \mathbf{A})$\;
%                \For{$h=H-1$ to $1$} {
%                     Compute $\mathbf{Z}^h$, $\mathbf{A}^h$, $\mathbf{C}^h$ with Eqs. \ref{assignment}, \ref{parentNode} and \ref{structure}\;
%                     Compute assignment $s^h_{ij}$ for $h=1, \cdots, H$ in Eq. \ref{eq.s}\;
%                }
%               Compute the objective of DSI in Eq. \ref{kse}\;
%               Optimize parameters via Riemannian Adam\;
%             }
%             Create a root node $\lambda$ and a queue $\mathcal{Q}$\;
%             \While{$\mathcal{Q}$ is not empty}{
%                 Get first item $\alpha$ in $\mathcal{Q}$\;
%                  Let $h=\alpha.h+1$ and search subsets $P$ from $[s^h_{ij}]$\;
%                  Create nodes from $P$ and put into the queue $\mathcal{Q}$\;
%                  Add theses nodes into $\alpha$'s children list\;
%             }
%             Return the partitioning tree $\mathcal{T}:=\lambda$\;
% \end{algorithm}

\paragraph{Complexity Analysis}
First, DSI loss is computed level-wisely, and for each level $h$ it has the time complexity of $O(N_h\lvert \mathcal{E} \rvert)$.
Second, the complexity of constructing the partitioning tree is $O(HN)$ with the breadth-first search in hyperbolic space. 
Thus, the overall complexity is yielded as $O(\lvert \mathcal{V} \rvert \lvert \mathcal{E} \rvert)$ with a small constant $H$, and $N_h \ll \lvert \mathcal{V} \rvert$ in intermediate layers.
Though the predefined $K$ is not required, it still presents similar complexity to most clustering models, i.e., $O(K\lvert \mathcal{V} \rvert^2)$, given real graphs are typically sparse.

% For the DSE loss complexity, From Equation \ref{kse}, our differentiable structural entropy can be computed level-wisely. For each level $h$, we have a $\mathcal{O}(N_h\lvert \mathcal{E} \rvert)$

% We construct the partitioning tree in the BFS manner, which only has a computational complexity of $\mathcal{O}(KN)$.

\begin{table*}[t]
  \centering
  \caption{Clustering results on Cora, Citerseer, AMAP, and Computer datasets in terms of NMI and ARI (\%). OOM denotes Out-of-Memory on our hardware. The best results are highlighted in \textbf{boldface}, and runner-ups are \underline{underlined}.}
    \vspace{0.05in}
  \label{tab.cluster}
    \resizebox{1\linewidth}{!}{
     \begin{tabular}{ c |cc|cc|cc|cc}
      \hline
         & \multicolumn{2}{c}{\textbf{Cora}}  & \multicolumn{2}{|c}{\textbf{Citeseer}}  & \multicolumn{2}{|c}{\textbf{AMAP}} & \multicolumn{2}{|c}{\textbf{Computer}} \\
        & NMI  & ARI  & NMI  & ARI  & NMI  & ARI  & NMI  & ARI \\
      \hline
       % Spectral  
       % & 29.70\scriptsize{$\pm$2.25}  & 17.40\scriptsize{$\pm$2.24}  &  \hphantom{0 }9.21\scriptsize{$\pm$3.01}  & \hphantom{0 }8.20\scriptsize{$\pm$3.14 }
       % & \hphantom{0 }0.60\scriptsize{$\pm$0.12}  & \hphantom{0 }0.03\scriptsize{$\pm$0.24}  & 00.00\scriptsize{$\pm$3.54}  & \hphantom{0 }6.40\scriptsize{$\pm$2.98} \\
      K-Means   
       & 14.98\scriptsize{$\pm$0.82}  & $\ \ $8.60\scriptsize{$\pm$0.40}  & 16.94\scriptsize{$\pm$0.24}  & 13.43\scriptsize{$\pm$0.57} 
       & 19.31\scriptsize{$\pm$3.75}  & 12.61\scriptsize{$\pm$3.54}  & 16.64\scriptsize{$\pm$0.75}  & $\ \ $2.71\scriptsize{$\pm$0.82} \\
      \hline
       VGAE \footnotesize{\cite{kipf2016variational}}
       & 43.40\scriptsize{$\pm$1.62}  & 37.50\scriptsize{$\pm$2.13}  & 32.70\scriptsize{$\pm$0.30}  & 33.10\scriptsize{$\pm$0.55}
       & 66.01\scriptsize{$\pm$0.84}  & 56.24\scriptsize{$\pm$0.22}  & 37.62\scriptsize{$\pm$0.24}  & 22.16\scriptsize{$\pm$0.16}\\
       ARGA \footnotesize{\cite{pan2018adversarially}}
       & 48.10\scriptsize{$\pm$0.45}  & 44.10\scriptsize{$\pm$0.28}  & 35.10\scriptsize{$\pm$0.58}  & 34.60\scriptsize{$\pm$0.48}
       & 58.36\scriptsize{$\pm$1.02}  & 44.18\scriptsize{$\pm$0.85}  & 37.21\scriptsize{$\pm$0.58}  & 26.28\scriptsize{$\pm$1.38}\\
       MVGRL \footnotesize{\cite{HassaniA20}}
       & 62.91\scriptsize{$\pm$0.42}  & 56.96\scriptsize{$\pm$0.74}  & 46.96\scriptsize{$\pm$0.25}  & 44.97\scriptsize{$\pm$0.57}
       & 36.89\scriptsize{$\pm$1.31}  & 18.77\scriptsize{$\pm$1.54}  & 10.12\scriptsize{$\pm$2.21}  & $\ \ $5.53\scriptsize{$\pm$1.78}\\
      Sublime \footnotesize{\cite{liu2022towards}}
       & 54.20\scriptsize{$\pm$0.28}  & 50.32\scriptsize{$\pm$0.32}  & 44.10\scriptsize{$\pm$0.27}  & 43.91\scriptsize{$\pm$0.35}
       & $\ \ $6.37\scriptsize{$\pm$0.54}  &$\ \ $5.36\scriptsize{$\pm$0.24}  & 39.16\scriptsize{$\pm$1.82}  & 24.15\scriptsize{$\pm$0.63}\\
       ProGCL \footnotesize{\cite{xia2021progcl}}
       & 41.02\scriptsize{$\pm$1.64}  & 30.71\scriptsize{$\pm$1.38}  & 39.59\scriptsize{$\pm$1.58}  & 36.16\scriptsize{$\pm$2.21}
       & 39.56\scriptsize{$\pm$0.25}  & 34.18\scriptsize{$\pm$0.58}  & 35.50\scriptsize{$\pm$2.06}  & 26.08\scriptsize{$\pm$1.91} \\
      \hline
      GRACE \footnotesize{\cite{yang2017graph}}
       & 57.30\scriptsize{$\pm$0.86}  & 52.70\scriptsize{$\pm$1.20}  & 39.90\scriptsize{$\pm$2.26}  & 37.70\scriptsize{$\pm$1.35}
       & 53.46\scriptsize{$\pm$1.32}  & 42.74\scriptsize{$\pm$1.57}  & 47.90\scriptsize{$\pm$1.65}  & 36.40\scriptsize{$\pm$1.56}\\
      DEC  \footnotesize{\cite{xie2016unsupervised}}
       & 23.54\scriptsize{$\pm$1.13}  & 15.13\scriptsize{$\pm$0.72}  & 28.34\scriptsize{$\pm$0.42}  & 28.12\scriptsize{$\pm$0.24}
       & 37.35\scriptsize{$\pm$0.84}  & 18.29\scriptsize{$\pm$0.64}  & 38.56\scriptsize{$\pm$0.24}  & 34.76\scriptsize{$\pm$0.35}\\
      % DCN (ICML17)  
      %  & 25.65\scriptsize{$\pm$1.45}  & 21.63\scriptsize{$\pm$0.54}  & 27.64\scriptsize{$\pm$0.44}  & 29.31\scriptsize{$\pm$0.24 }
      %  & 31.46\scriptsize{$\pm$2.03}  & 15.70\scriptsize{$\pm$1.18}  & 00.00\scriptsize{$\pm$0.00}  & 00.00\scriptsize{$\pm$0.00 }\\
      MCGC  \footnotesize{\cite{pan2021multi}}
       & 44.90\scriptsize{$\pm$1.56}  & 37.80\scriptsize{$\pm$1.24}  & 40.14\scriptsize{$\pm$1.44}  & 38.00\scriptsize{$\pm$0.85}
       & 61.54\scriptsize{$\pm$0.29}  & 52.10\scriptsize{$\pm$0.27}  & 53.17\scriptsize{$\pm$1.29}  & 39.02\scriptsize{$\pm$0.53}\\
      DCRN \footnotesize{\cite{liu2021deep}}
       & 48.86\scriptsize{$\pm$0.85}  & 43.79\scriptsize{$\pm$0.48}  & 45.86\scriptsize{$\pm$0.35}  & \underline{47.64}\scriptsize{$\pm$0.30}
       & \underline{73.70}\scriptsize{$\pm$1.24}  & 63.69\scriptsize{$\pm$0.84}  & OOM                                       & OOM      \\
       %& 12.61\scriptsize{$\pm$0.57}  & \hphantom{0 }4.82\scriptsize{$\pm$0.26 }\\
       FT-VGAE  \footnotesize{\cite{DBLP:conf/ijcai/MrabahBK22}}
        & 61.03\scriptsize{$\pm$0.52}   & 58.22\scriptsize{$\pm$1.27}   & 44.50\scriptsize{$\pm$0.13}  & 46.71\scriptsize{$\pm$0.75} 
        & 69.76\scriptsize{$\pm$1.06}   & 59.30\scriptsize{$\pm$0.81}   & 51.36\scriptsize{$\pm$0.92}  & \underline{40.07}\scriptsize{$\pm$2.13} \\
       gCooL  \footnotesize{\cite{DBLP:conf/www/LiJT22}}
       & 58.33\scriptsize{$\pm$0.24}   & 56.87\scriptsize{$\pm$1.03}  & 47.29\scriptsize{$\pm$0.10}  & 46.78\scriptsize{$\pm$1.51} 
       & 63.21\scriptsize{$\pm$0.09}   & 52.40\scriptsize{$\pm$0.11}  & 47.42\scriptsize{$\pm$1.76}  & 27.71\scriptsize{$\pm$2.28}\\
       S$^3$GC  \footnotesize{\cite{devvrit2022s3gc}}
       & 58.90\scriptsize{$\pm$1.81}  & 54.40\scriptsize{$\pm$2.52}  & 44.12\scriptsize{$\pm$0.90}  & 44.80\scriptsize{$\pm$0.65}
       & 59.78\scriptsize{$\pm$0.45}  & 56.13\scriptsize{$\pm$0.58}  & \underline{54.80}\scriptsize{$\pm$1.22}  & 29.93\scriptsize{$\pm$0.22}\\
       Congregate  \footnotesize{\cite{sun2023contrastive}}
       & \underline{63.16}\scriptsize{$\pm$0.71}  & 59.27\scriptsize{$\pm$1.23}  & \underline{50.92}\scriptsize{$\pm$1.58}  & 47.59\scriptsize{$\pm$1.60} 
       & 70.99\scriptsize{$\pm$0.67}  & 60.55\scriptsize{$\pm$1.36}  & 46.03\scriptsize{$\pm$0.47}  & 38.57\scriptsize{$\pm$1.05}\\
      DinkNet  \footnotesize{\cite{liu2023dink}}
       & 62.28\scriptsize{$\pm$0.24}  & 61.61\scriptsize{$\pm$0.90}  & 45.87\scriptsize{$\pm$0.24}  & 46.96\scriptsize{$\pm$0.30}
       & \textbf{74.36}\scriptsize{$\pm$1.24}  & \textbf{68.40}\scriptsize{$\pm$1.37}  & 39.54\scriptsize{$\pm$1.52} & 33.87\scriptsize{$\pm$0.12}\\
      GC-Flow  \footnotesize{\cite{wang2023gc}}
       & 62.15\scriptsize{$\pm$1.35}  & \underline{63.14}\scriptsize{$\pm$0.80}  & 40.50\scriptsize{$\pm$1.32}  & 42.62\scriptsize{$\pm$1.44}
       & 36.45\scriptsize{$\pm$1.21}  & 37.24\scriptsize{$\pm$1.26}  & 41.10\scriptsize{$\pm$1.06} & 35.60\scriptsize{$\pm$1.82 }\\
      \hline
       RGC \footnotesize{\cite{liu2023reinforcement}}
       & 57.60\scriptsize{$\pm$1.36}  & 50.46\scriptsize{$\pm$1.72}  & 45.70\scriptsize{$\pm$0.29}  & 45.47\scriptsize{$\pm$0.43}
       & 47.65\scriptsize{$\pm$0.91}  & 42.65\scriptsize{$\pm$1.53}  & 46.24\scriptsize{$\pm$1.05}  & 36.12\scriptsize{$\pm$0.30} \\
      \textbf{\texttt{LSEnet}} (Ours)
     & \textbf{63.97}\scriptsize{$\pm$0.67}  & \textbf{63.35}\scriptsize{$\pm$0.56}   & \textbf{52.26}\scriptsize{$\pm$1.09}  &\textbf{48.01}\scriptsize{$\pm$1.25}  
       & 71.72\scriptsize{$\pm$1.30}  & \underline{65.08}\scriptsize{$\pm$0.73}  & \textbf{55.03}\scriptsize{$\pm$0.79} & \textbf{42.15}\scriptsize{$\pm$1.02}\\
      \hline
    \end{tabular}
}
\end{table*}

\subsection{Hyperbolic Partitioning Tree}

% In the theory of structural entropy, the optimal partitioning tree of the graph is constructed in the principle of structural entropy minimization \cite{li2016structural}.
In the principle of structural information minimization, the optimal partitioning tree is constructed to describe the self-organization of the graph \cite{li2016structural,wu2023sega}.
In the continuous realm, 
%with Theorem \ref{theorem.equivalence} (Equivalence), 
\emph{\texttt{LSEnet} learns the optimal partitioning tree in hyperbolic space} by minimizing the objective in Eq. (\ref{kse}).
DSI is applied on all the level-wise assignment $\mathbf C$'s, which are parameterized by neural networks in Sec. \ref{leaf} and \ref{parent}.
%we level-wisely build the partitioning tree  from bottom to top.
We suggest to place the tree root at the origin of $\mathbb L^{\kappa, d_{\mathcal T}}$, so that the learnt tree enjoys the symmetry of Lorentz model.
Consequently, the hyperbolic partitioning tree describes the graph's self-organization and clustering structure in light of Theorem \ref{theorem.conductance}.
% Concretely, we enforce the regularization as follows,
% \begin{align}
%     \mathcal L_{root}= d_{\mathbb L}(\boldsymbol z^{0},  \boldsymbol o),
%     \label{loss.root}
% \end{align}
% where the origin is $...$

% The main algorithm to decode a partitioning tree from leaves embeddings $\mathbf{Z}$ in hyperbolic space, details are shown in Algorithm \ref{alg:decode} given in Appendix \ref{append.alg}. The partitioning tree is constructed from top to bottom, opposite to the process in \texttt{LSEnet}. This reverse-construction manner can make a relaxed partitioning tree into a strict one, such as vanishing the empty leaf nodes and avoiding a chain appearing in the tree, as the process performs.

%%%%%%%%%%%%%%%%%%%%%%%%%%%%%%%%%%%%%%%%%%%%%%%%%%%%%%%%%%%

The overall procedure of training \texttt{LSEnet} is summarized in Algorithm \ref{alg:lsenet}.
Specifically, in Lines $1$-$9$, we learn level-wise assignments and node embeddings in a bottom-up manner.
In Lines $10$-$17$, the optimal partitioning tree is given via a top-down process in which we avoid empty leaf nodes and fix single chains in the relaxed tree (case one and case two in Theorem \ref{theorem.flex}).
% In the bottom-up construction process, we can not avoid empty leaf nodes. Moreover, a chain may appear in the relaxed partitioning tree, like case 2 in Theorem \ref{theorem.flex}, because there may be a level such that all points belong to exactly one node, which will not impact structural information. To fix these relaxations, we propose a decoding algorithm.
Further details of the procedure are shown in Algorithm \ref{alg:decode} given in Appendix \ref{append.alg}.

\emph{In brief, \texttt{LSEnet} combines the advantages of both structural entropy and hyperbolic space, and uncovers the clustering structures without  predefined cluster number $K$.}

\section{Experiments}
We conduct extensive experiments on benchmark datasets to evaluate the effectiveness of the proposed \texttt{LSEnet}.\footnote{Datasets and codes of  \texttt{LSEnet} are available at the anonymous link \url{https://anonymous.4open.science/r/LSEnet-3EF2}} Furthermore, we compare with the classic formulation, evaluate the parameter sensitivity, and visualize the hyperbolic portioning tree.
Additional results are shown in Appendix D.

\subsection{Graph Clustering}

\paragraph{Datasets \& Baselines.}
The evaluations are conducted on $4$ datasets, including Cora, Citeseer, and Amazon Photo (AMAP) \cite{liu2023dink}, and a larger Computer dataset \cite{DBLP:conf/www/LiJT22}.
%\citeliu2023reinforcement}, and a larger XXX.
We focus on deep graph clustering in this paper, and thus we primarily compare with the deep models, including $11$ graph clustering methods, i.e., 
GRACE,
DEC,
MCGC,
DCRN ,
FT-VGAE,
gCooL,
S$^3$GC,
Congregate, 
RGC,
DinkNet, 
and GC-Flow,
and $5$ self-supervised GNNs, i.e., 
VGAE, 
ARGA,  
Sublime, 
ProGCL and  
MVGRL.
Additionally, we provide the results of $K$-Means as a reference.
Datasets and baselines are detailed in Appendix \ref{append.database}.

\paragraph{Metric \& Configuration.}
Both Normalized Mutual Information (NMI) and Adjusted Rand Index (ARI) \cite{HassaniA20,DBLP:conf/www/LiJT22} are employed as the evaluation metric for the clustering results. 
In \texttt{LSEnet}, we work in the Lorentz model with curvature  $\kappa=-1$. The $\operatorname{MLP}$ of  Lorentz assigner  has $3$ layers. The dimension of structural entropy is a hyperparameter. 
The parameters are optimized by Riemannian Adam \cite{GeoOpt} with a $0.003$ learning rate.
For all the methods, the cluster number $K$ is set as the number of ground-truth classes,
and we report the mean value with standard deviations of $10$ independent runs.
Further details are in Appendix \ref{append.note}.

\begin{figure}[t]
\centering 
\subfigure[NMI]{
\includegraphics[width=0.48\linewidth]{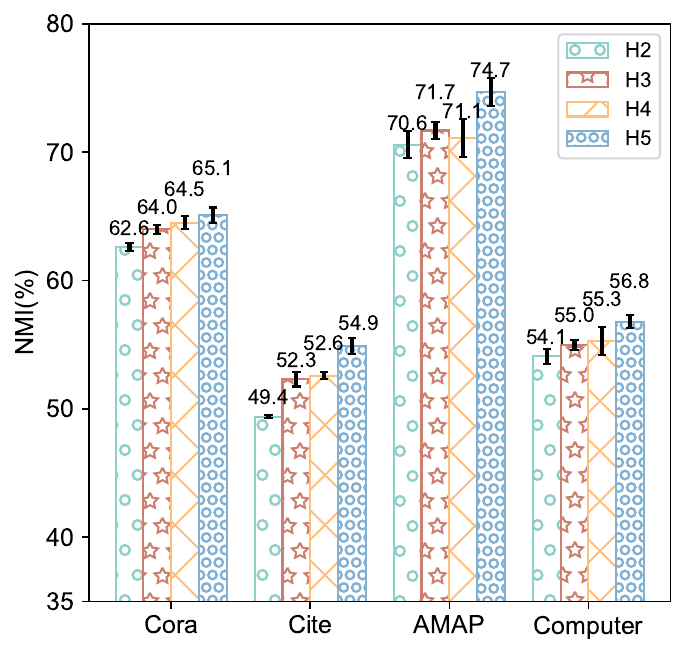}}
  \hspace{-0.05in}
\subfigure[ARI]{
\includegraphics[width=0.48\linewidth]{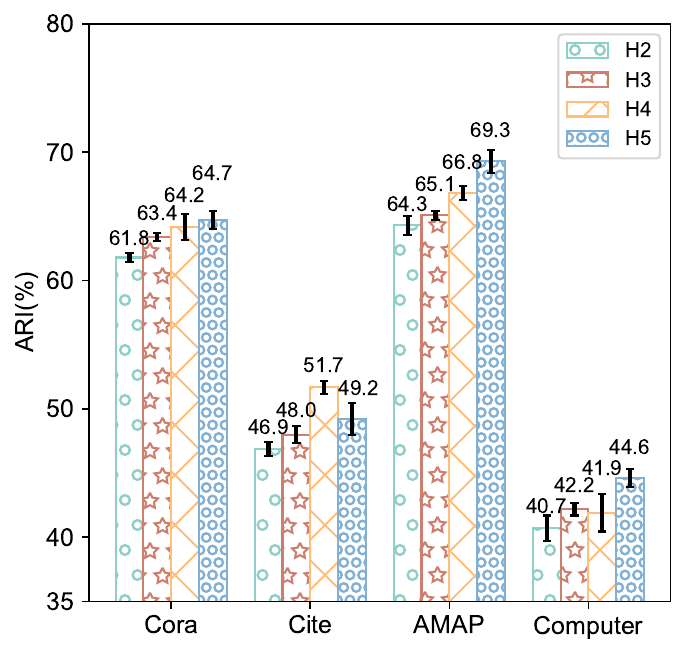}}
  \vspace{-0.15in}
\caption{Parameter sensitivity on dimension of structural entropy.}
\label{fig:sensitive}
  \vspace{-0.1in}
\end{figure}

\begin{table*}[t]
  \centering
  \caption{Comparison between \texttt{LSEnet} and CSE on Cora, Citerseer, Football and Karate datasets in terms of NMI and ARI (\%). }
  \vspace{0.05in}
  \label{tab.compare}
    \resizebox{0.9\linewidth}{!}{
     \begin{tabular}{ c |cc|cc|cc|cc}
      \hline
         & \multicolumn{2}{c}{\textbf{Cora}}  & \multicolumn{2}{|c}{\textbf{Citeseer}}  & \multicolumn{2}{|c}{\textbf{Football}} & \multicolumn{2}{|c}{\textbf{Karate}} \\
        & NMI  & ARI  & NMI  & ARI  & NMI  & ARI  & NMI  & ARI \\
      \hline
    CSE
       & 60.82\scriptsize{$\pm$0.30}  & 59.36\scriptsize{$\pm$0.17}  & 47.12\scriptsize{$\pm$1.01}  & 45.67\scriptsize{$\pm$0.67}
       & 79.23\scriptsize{$\pm$0.12}  & 54.06\scriptsize{$\pm$0.05}  & 81.92\scriptsize{$\pm$0.29}  & 69.77\scriptsize{$\pm$0.02} \\
       \textbf{\texttt{LSEnet}}
       & 62.57\scriptsize{$\pm$0.59}  & 61.80\scriptsize{$\pm$0.67}  & 49.35\scriptsize{$\pm$0.20}  & 46.91\scriptsize{$\pm$1.12} 
       & 80.37\scriptsize{$\pm$1.02}  & 54.72\scriptsize{$\pm$2.05}  & 82.19\scriptsize{$\pm$3.16}  & 70.31\scriptsize{$\pm$0.80} \\
      \hline
      \textbf{Performance Gain}
      & \textbf{1.75} & \textbf{2.44} &\textbf{2.23} &\textbf{1.24}
      & \textbf{1.14} & \textbf{0.66} &\textbf{0.27} &\textbf{0.54}\\
      \hline
    \end{tabular}
}
  \vspace{-0.1in}
\end{table*}

\paragraph{Comparison with State-of-the-art.} 
Clustering results on Cora, Citeseer, AMAP and Computer datasets are collected in Table \ref{tab.cluster}.
Concretely, self-supervised GNNs do not generate node clusters themselves, and we apply $K$-means to obtain the results.
 In \texttt{LSEnet}, node features are projected to Lorentz model via the exponential map (Eq. \ref{expmap}) with respect to the origin point.
 \texttt{LSEnet} learns the hyperbolic partitioning tree, and node clusters are given in a divisive manner with hyperbolic distance, detailed as Algorithm 3 in Appendix \ref{append.alg}.
 As shown in Table \ref{tab.cluster}, even though all the baselines except RGC have the cluster number $K$ as input parameter,
 %the recent contrastive methods tend to obtain better results
the proposed  \texttt{LSEnet} achieves the best performance except on AMAP dataset and has more consistent performance, e.g., DinkNet does well on AMAP but worse on Computer and Citeseer.
Without $K$, RGC automatically seeks $K$ via reinforcement learning, while we present another idea of learning a partitioning tree and have better results, e.g., over 20\% NMI gain on AMAP dataset.
\texttt{LSEnet} takes advantage of structural entropy, encoding the self-organization of graphs, and hyperbolic geometry further benefits graph clustering, as shown in the next Sec.

% \begin{table}[t]
%   \vspace{-0.05in}
%   \centering
%   \caption{Embedding expressiveness regarding link prediction on Cora, Citeseer, AMAP and Computer datasets.}
%   \vspace{0.05in}
%   \label{tab.link}
%    \resizebox{1\linewidth}{!}{
%      \begin{tabular}{ c |c c c c}
%       \hline
%                  & {\textbf{Cora}}  & {\textbf{Citeseer}}  & {\textbf{AMAP}} & {\textbf{Computer}}\\
%      \hline
%         GCN
%        & 90.11\scriptsize{$\pm$0.51}  & 90.16\scriptsize{$\pm$0.49}  & 00.00\scriptsize{$\pm$3.54}  & 00.00\scriptsize{$\pm$3.54}  \\
%         Sage
%        & 86.02\scriptsize{$\pm$0.55}  & 88.18\scriptsize{$\pm$0.22}  & 00.00\scriptsize{$\pm$3.54}  & 00.00\scriptsize{$\pm$3.54}  \\
%         DGI
%        & 92.6 \scriptsize{$\pm$0.02}  & 93.3\scriptsize{$\pm$0.04}  & 00.00\scriptsize{$\pm$3.54}  & 00.00\scriptsize{$\pm$3.54}  \\
%       \hline 
%        HGCN
%        & 93.60\scriptsize{$\pm$0.37}  & 94.33\scriptsize{$\pm$0.42}  & 00.00\scriptsize{$\pm$3.54}  & 9  6.88\scriptsize{$\pm$0.53}  \\
%        LGCN
%        & 92.69 \scriptsize{$\pm$0.26}  & 93.49\scriptsize{$\pm$1.11}  & 00.00\scriptsize{$\pm$3.54}  & 96.37\scriptsize{$\pm$0.70}  \\
%        $Q$GCN
%        & 95.22 \scriptsize{$\pm$0.29}  & 94.31\scriptsize{$\pm$0.73}  & 00.00\scriptsize{$\pm$3.54}  & 95.1\scriptsize{$\pm$0.03}  \\
%       \texttt{LSEnet}
%        & 13.23\scriptsize{$\pm$1.56}  & 5.5\scriptsize{$\pm$2.14}  & 16.64\scriptsize{$\pm$0.75}  & 00.00\scriptsize{$\pm$3.54}  \\
%       \hline
%     \end{tabular}
% }
%   \vspace{-0.2in}
% \end{table}

\begin{table}[t]
  \centering
  \caption{Embedding expressiveness regarding link prediction on Cora, Citeseer, AMAP and Computer datasets.}
  \vspace{0.05in}
  \label{tab.link}
   \resizebox{1\linewidth}{!}{
     \begin{tabular}{ c |c c c c}
      \hline
                 & {\textbf{Cora}}  & {\textbf{Citeseer}}  & {\textbf{AMAP}} & {\textbf{Computer}}\\
     \hline
        GCN
       & 91.19\scriptsize{$\pm$0.51}  & 90.16\scriptsize{$\pm$0.49}  & 90.12\scriptsize{$\pm$0.72}  &  93.86\scriptsize{$\pm$0.36}  \\
        SAGE
       & 86.02\scriptsize{$\pm$0.55}  & 88.18\scriptsize{$\pm$0.22}  & 98.02\scriptsize{$\pm$0.32}  &  92.82\scriptsize{$\pm$0.20}  \\
        GAT
       & 92.55\scriptsize{$\pm$0.49}  & 89.32\scriptsize{$\pm$0.36}  & \underline{98.67}\scriptsize{$\pm$0.08}  &  95.93\scriptsize{$\pm$0.19} \\
      \hline 
       HGCN
       & 93.60\scriptsize{$\pm$0.37}  & \textbf{94.39}\scriptsize{$\pm$0.42}  & 98.06\scriptsize{$\pm$0.29}  & \underline{96.88}\scriptsize{$\pm$0.53}  \\
       LGCN
       & 92.69\scriptsize{$\pm$0.26}  & 93.49\scriptsize{$\pm$1.11}  & 97.08\scriptsize{$\pm$0.08}  & 96.37\scriptsize{$\pm$0.70}  \\
       $Q$GCN
       & \underline{95.22}\scriptsize{$\pm$0.29}  & 94.31\scriptsize{$\pm$0.73}  & 95.17\scriptsize{$\pm$0.45} & 95.10\scriptsize{$\pm$0.03}  \\
       \hline 
     \textbf{ \texttt{LSEnet}}
       & \textbf{95.51}\scriptsize{$\pm$0.60}  & \underline{94.32}\scriptsize{$\pm$1.51}  & \textbf{98.75}\scriptsize{$\pm$0.67}  &  \textbf{97.06}\scriptsize{$\pm$1.02} \\
      \hline
    \end{tabular}
}
  \vspace{-0.1in}
\end{table}

\subsection{Discussion on Structural Entropy}

\paragraph{Parameter Sensitivity}
%To verify how the dimensional number influences the performance of our model, we conduct experiments on four datasets under different numbers of 

We examine the parameter sensitivity on the dimension of structural entropy in \texttt{LSEnet}, i.e., the height of the partitioning tree $H$. 
The clustering results in Table \ref{tab.cluster} is given by $H=3$. Here, we vary the height $H$ in $\{2, 3, 4, 5\}$, and report the results on Cora, Citeseer, AMAP and Computer datasets in Fig. \ref{fig:sensitive}. 
It shows that \texttt{LSEnet} generally receives performance gain when increasing the height. Also, \texttt{LSEnet} is able to obtain satisfactory results with small heights.

% We can find that as the dimension grows, the variance increases as well, since constructing a partitioning tree is an NP-hard problem that requires exponential search space when the height of the tree increases, the gradient descent methods can not guarantee the optimal solution.

\begin{figure}[t]
\centering 
\subfigure[Cora]{
\includegraphics[scale=0.17]{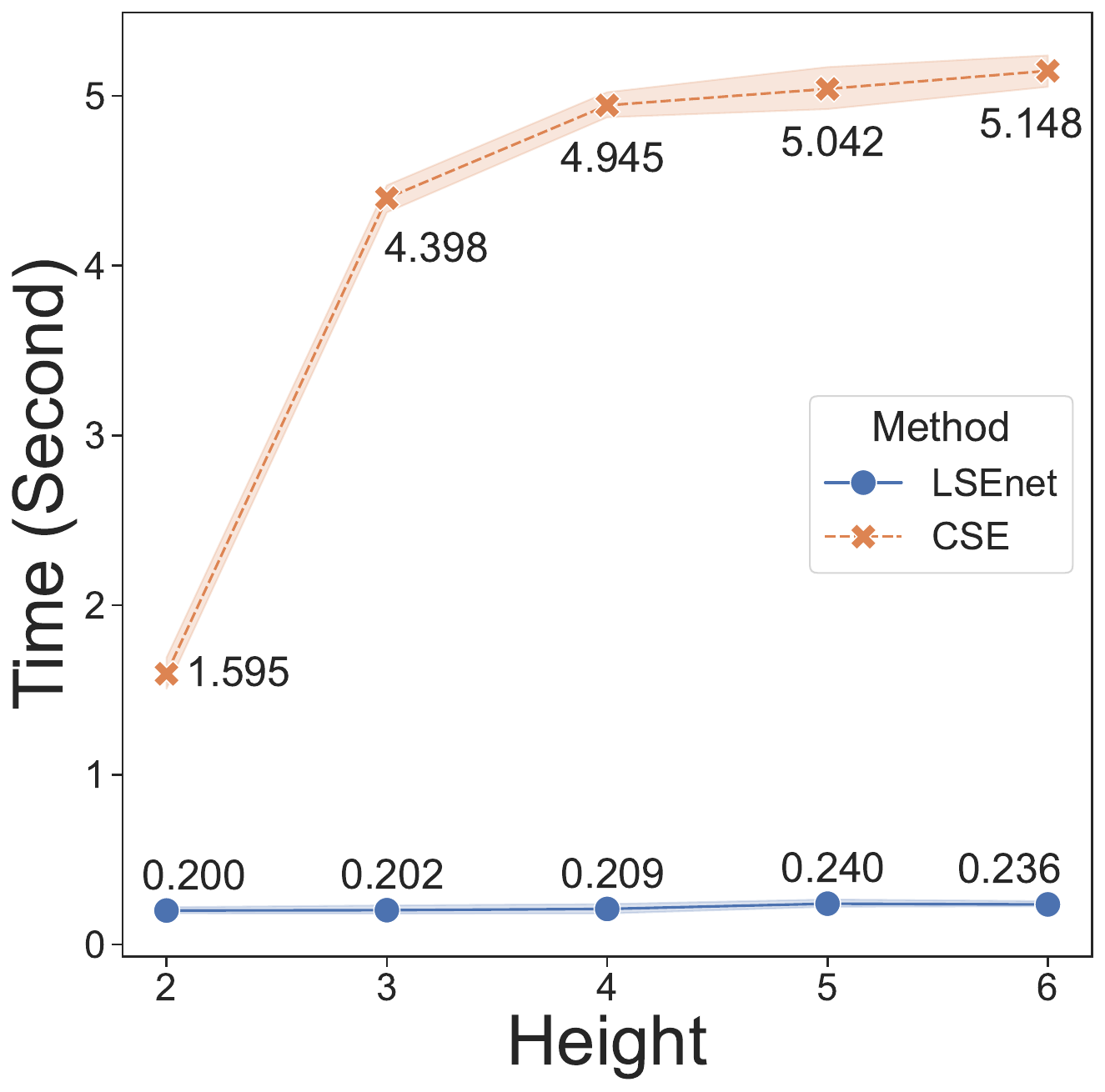}}
\subfigure[Football]{
\includegraphics[scale=0.17]{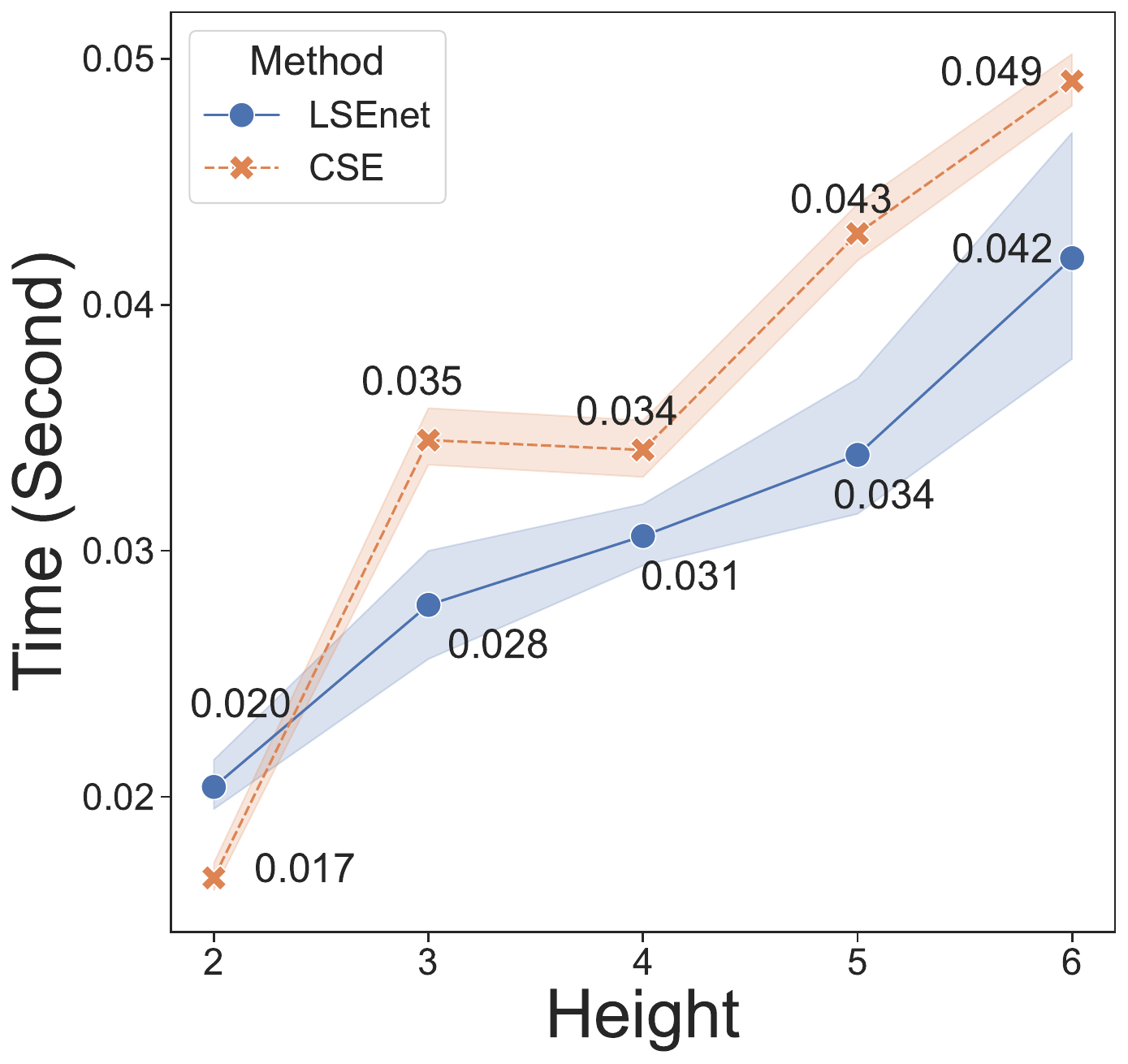}}
\caption{Running time to obtain partitioning tree}
\label{running_time}
  \vspace{-0.1in}
\end{figure}

\paragraph{Comparison with Classic Structural Entropy}
%We compare with the classic structural entropy \citet{li2016structural}, termed as CSE.
% in terms of complexity and effectiveness.
% First, on time complexity, $\mathcal{O}(H(\lvert \mathcal{V} \rvert + \lvert \mathcal{E} \rvert))$ of \texttt{LSEnet} scales linearly with the graph size $\lvert \mathcal{V} \rvert$ and the dimension $H$,
First, the time complexity of \texttt{LSEnet} is $O(\lvert \mathcal{V} \rvert \lvert \mathcal{E} \rvert)$ regardless of the dimension of structural information,
while the classic structural entropy \cite{li2016structural}, termed as CSE, scales exponentially with respect to its dimension (i.e., tree height).
% e.g., $\mathcal{O}(\lvert \mathcal{V} \rvert^3 \log^2 \lvert \mathcal{V} \rvert)$ for the $3$-dimensional case, and it is unacceptable for large graphs.
For example, it is $O(\lvert \mathcal{V} \rvert^3 \log^2 \lvert \mathcal{V} \rvert)$ for the $3$-dimensional case, which is unacceptable for large graphs.
Second, on running time, Fig. \ref{running_time} shows the running time to obtain the partitioning tree of different heights. 
%CSE presents competitive time to \texttt{LSEnet} on Football, but has at least $7\times$ time cost on larger Cora.
CSE has a competitive time cost to \texttt{LSEnet} on the small dataset (Football), but does badly on larger datasets, e.g., at least $7 \times$ time cost to \texttt{LSEnet} on Cora.
Third, we examine the effectiveness of clustering. 
CSE itself is unaware of node features, and cannot perform clustering. Instead, we consider the $2$-dimensional case and treat the edges between leaves (graph nodes) and layer-one nodes (regarded as clusters) as clustering results.
%Empirical results on Cora and Citeseer are reported in Table 2.
For a fair comparison, same as CSE, we set $H$ as $2$ in \texttt{LSEnet}  and consider the assignment as the result.
We report the results on Cora, Citeseer, Football and Karate in Table \ref{tab.compare}. \texttt{LSEnet} achieves  better results than CSE. 
It suggests that \emph{hyperbolic space of \texttt{LSEnet} benefits node-cluster assignment.}

%Clustering results are induced from the partitioning tree via Algo. 3, and 

\paragraph{Embedding Expressiveness.}
In addition to the cluster assignment, \texttt{LSEnet} learns node embeddings in hyperbolic space.
We evaluate the embedding expressiveness regarding \emph{link prediction}.
We compare with the popular GCN \cite{kipf2017semisupervised}, SAGE \cite{hamilton2018inductive} and GAT \cite{velickovic2018graph} in Euclidean space, 
hyperbolic models including HGCN \cite{chami2019hyperbolica} and LGCN \cite{zhang2021lorentzian}, 
and a recent $Q$GCN in ultra hyperbolic space \cite{DBLP:conf/nips/XiongZPP0S22}.
Results in terms of AUC are provided in Table \ref{tab.link}, where we set $H$ as $3$ for \texttt{LSEnet}.
%\texttt{LSEnet} presents superiority to the baselines, 
It shows that hyperbolic embeddings of \texttt{LSEnet}  encode the structural information for link prediction.
% as well.
%demonstrating the expressiveness

\begin{figure}[t]
\centering 
\subfigure[Karate: Results]{
\includegraphics[width=0.48\linewidth]{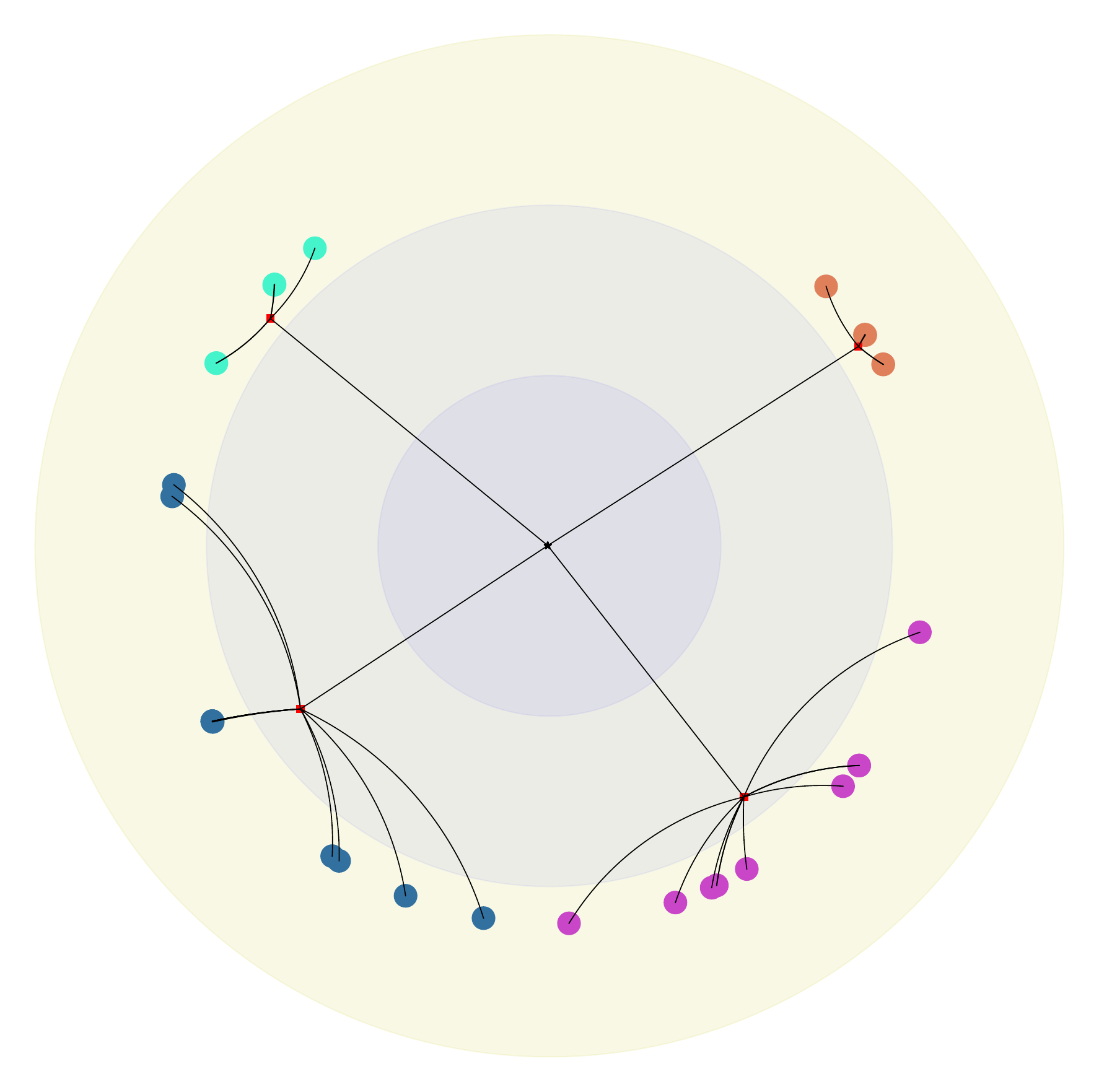}}
\subfigure[Football: Results]{
\includegraphics[width=0.48\linewidth]{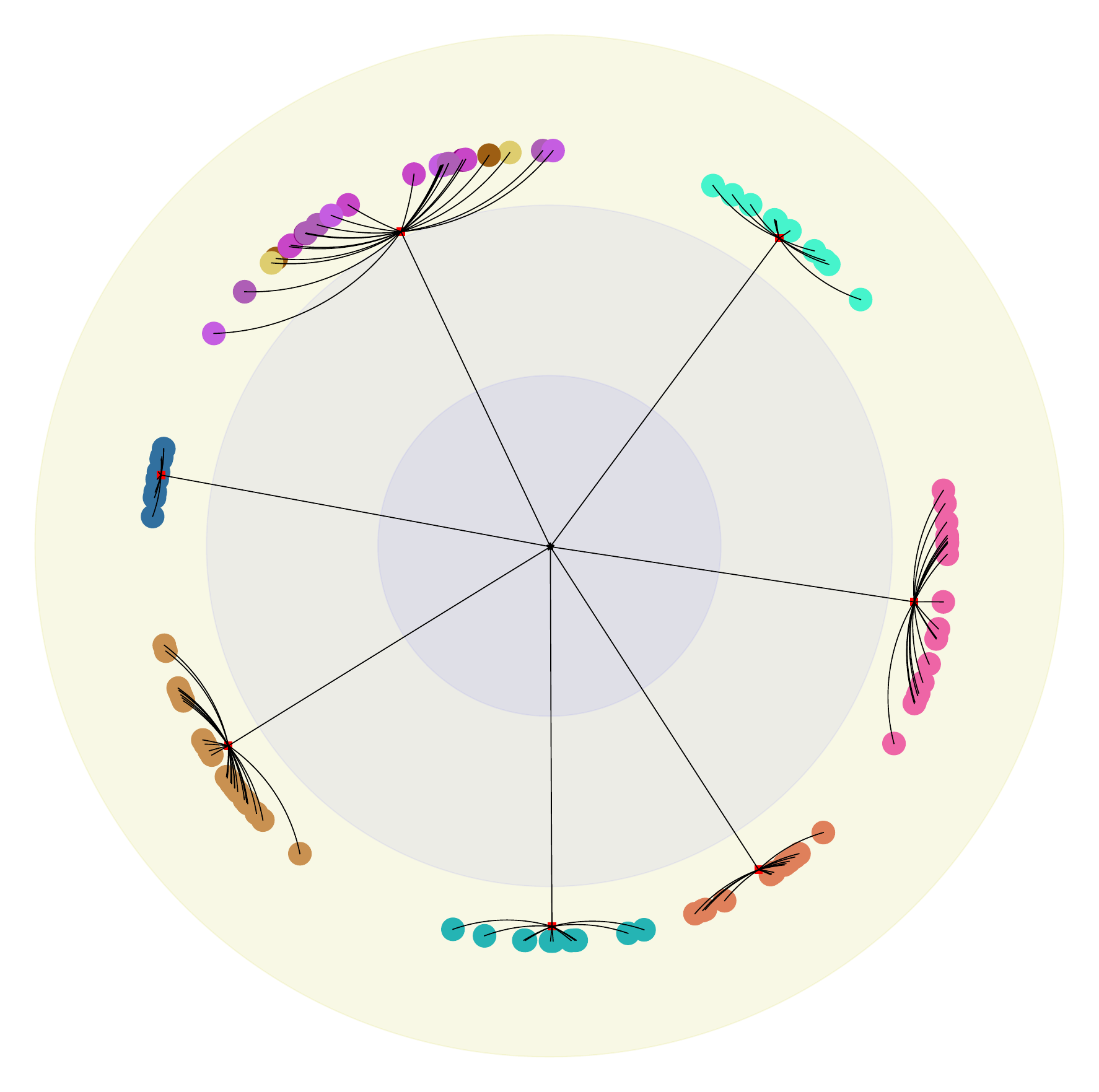}}
\subfigure[Karate: Groundtruth]{
\includegraphics[width=0.48\linewidth]{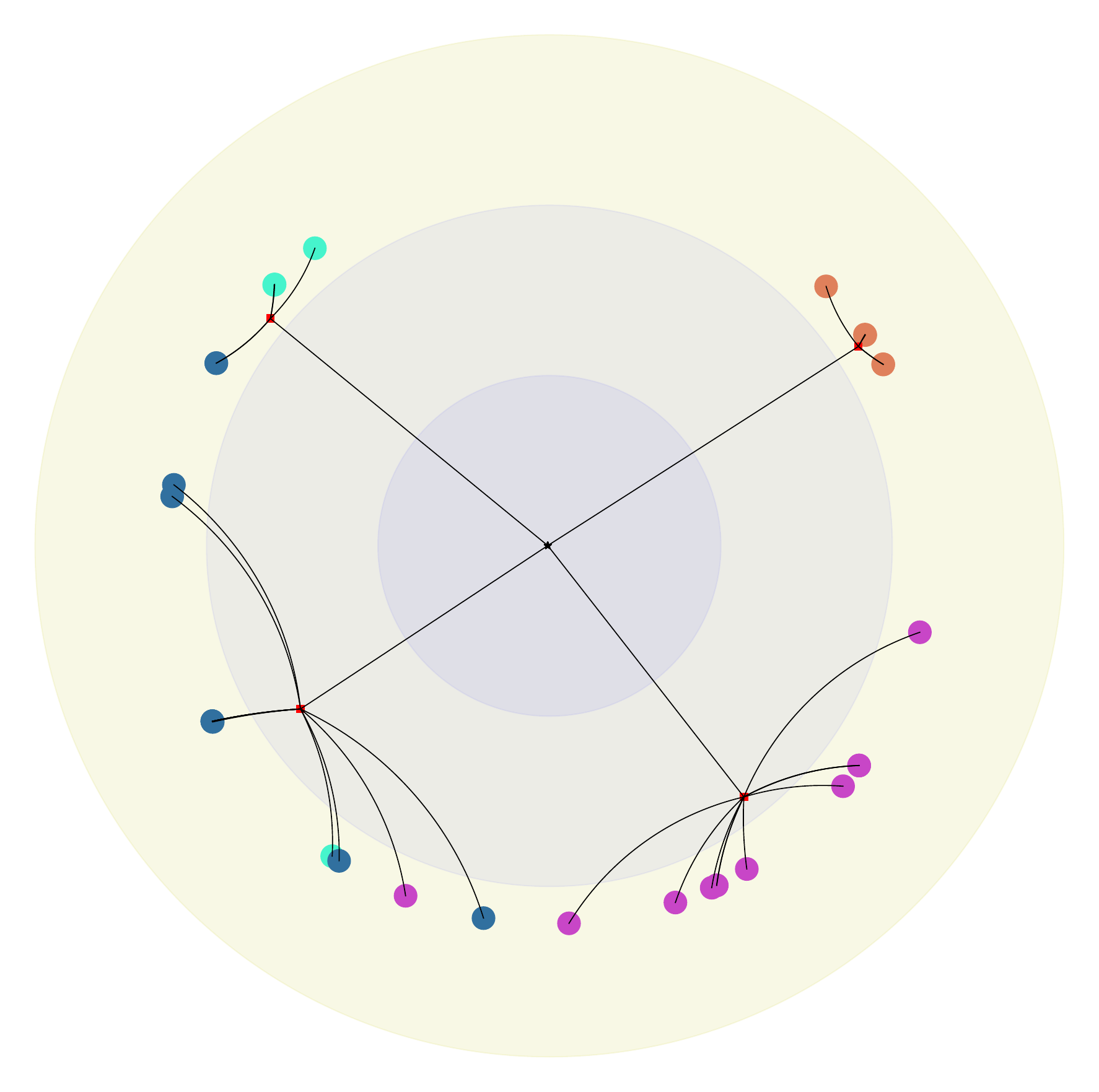}}
\subfigure[Football: Groundtruth]{
\includegraphics[width=0.48\linewidth]{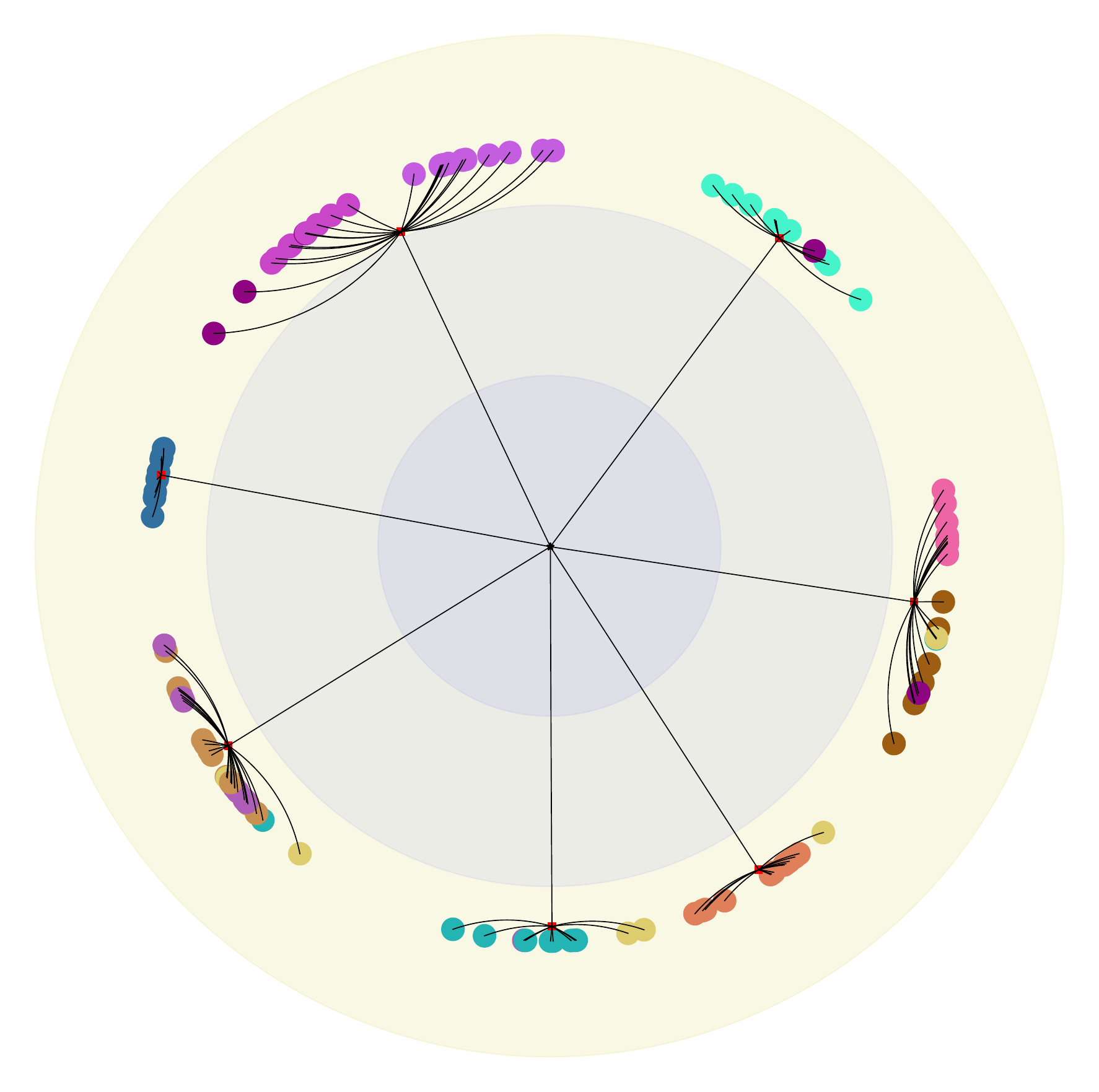}}
\caption{Visualization of hyperbolic partitioning trees.}
\label{vis}
  \vspace{-0.1in}
\end{figure}

\paragraph{Case Study \& Visualization}
%In the case study, 
Here, we visualize hyperbolic partitioning trees of \texttt{LSEnet}, and discuss graph clustering on Football and Karate datasets.
%Note that, the $2$-dimensional Lorentz model of \texttt{LSEnet} is represented in the $3$-dimensional Euclidean space, and is hard to visualize.
In \texttt{LSEnet}, the partitioning tree lives in the $2$-dimensional Lorentz model, which is represented in the $3$-dimensional Euclidean space and hard to visualize.
We conduct stereographic projection $\Psi$ \cite{bachmann2020constant} on the Lorentz model and obtain the corresponding Poincar\'{e} disc $\mathbb{B}^2$, which is more preferable to visualization.
Specifically, with the curvature $\kappa=-1$, for any $\boldsymbol x=[x_1 \ \boldsymbol x_s]^\top \in \mathbb L^2$, we have $\Psi(\boldsymbol{x})=\frac{\boldsymbol x_s}{x_1 + 1} \in \mathbb{B}^2$.
We plot the learned clusters in Fig. \ref{vis} (a) and (b), while ground-truth classes are in Fig. \ref{vis} (c) and (d), where the clusters are denoted by different colors.
%It shows that our results are well aligned with the ground-truth class labels.
The learnt clusters are well separated, and are aligned with the real clusters.
The visualization of Cora is in Appendix \ref{append.add}.

\section{Conclusion}

% In this paper, we rethink graph clustering from the perspective of graph information theory, and make the first attempt to remove the prior knowledge of cluster number.
% In particular, we establish the theory on differentiable structural entropy, and design a neural model (\texttt{LSEnet}) ...
% we integrate node features and structural entropy via a graph convolution in the manifold, 
% and learn the optimal encoding tree in the Lorentz model of hyperbolic space for graph clustering.
% Extensive experiments examine   show 
In this paper, we study graph clustering without predefined cluster number from a new perspective of structural entropy.
% We formulate a new objective of graph clustering, DSI,  not requiring the cluster number, and we present several theoretical results of DSI.
% Furthermore, we design a \texttt{LSEnet} with graph neural network in Lorentz model of hyperbolic space.
% \texttt{LSEnet}  learns the optimal partitioning tree of a graph by minimizing DSI.
We formulate a new graph clustering objective (DSI) not requiring the cluster number, so that the clusters are revealed in the optimal partitioning tree given by DSI minimization. Furthermore, we present a neural \texttt{LSEnet} to learn the optimal tree in hyperbolic space, where we integrate node features to structural information with Lorentz graph convolution net.
Extensive empirical results show the superiority of our approach on real graphs.

%\section*{Acknowledgements}

% \textbf{Do not} include acknowledgements in the initial version of
% the paper submitted for blind review.

% If a paper is accepted, the final camera-ready version can (and
% probably should) include acknowledgements. In this case, please
% place such acknowledgements in an unnumbered section at the
% end of the paper. Typically, this will include thanks to reviewers
% who gave useful comments, to colleagues who contributed to the ideas,
% and to funding agencies and corporate sponsors that provided financial


\section*{Broader Impact}
This paper presents work whose goal is to bridge the structural entropy of information theory and deep learning and shows a new objective function for deep graph clustering, not requiring the predefined cluster number. There are many potential societal consequences of our work, none of which we feel must be specifically highlighted here.

\nocite{langley00}

\bibliography{icml2024}
\bibliographystyle{icml2024}

%%%%%%%%%%%%%%%%%%%%%%%%%%%%%%%%%%%%%%%%%%%%%%%%%%%%%%%%%%%%%%%%%%%%%%%%%%%%%%%
%%%%%%%%%%%%%%%%%%%%%%%%%%%%%%%%%%%%%%%%%%%%%%%%%%%%%%%%%%%%%%%%%%%%%%%%%%%%%%%
% APPENDIX
%%%%%%%%%%%%%%%%%%%%%%%%%%%%%%%%%%%%%%%%%%%%%%%%%%%%%%%%%%%%%%%%%%%%%%%%%%%%%%%
%%%%%%%%%%%%%%%%%%%%%%%%%%%%%%%%%%%%%%%%%%%%%%%%%%%%%%%%%%%%%%%%%%%%%%%%%%%%%%%

%!TEX root = ./main.tex

\appendix
\onecolumn
\setcounter{equation}{0}

\section*{Appendix}

The appendix is structured in four sections.
\textbf{A. Proofs} on differential structural information,  
\textbf{B. Hyperbolic Space}, including important notions and facts in hyperbolic geometry, 
 \textbf{C. Technical Details} on algorithms, datasets, baselines,  etc., and
\textbf{D. Additional Results} on real datasets.

\section{Proofs}
\label{proofs}
Here, we detail lemmas and theorems on the proposed \textbf{Differential Structural Information}. In particular, we prove the \emph{equivalence, additivity, flexibility, bound} and \emph{the connection to graph clustering}, and show the supporting lemmas. 
%, in which we give the lemmas not shown in the paper due to the limit of space.
%(Theorems in hyperbolic space are elaborated in the next section.)

\subsection{Proof of Theorem \ref{theorem.equivalence}}
\label{proof.equivalence}
\textbf{Theorem 4.4} (Equivalence)
\emph{The formula $\mathcal{H}^\mathcal{T}$ in Definition \ref{DSE} is equivalent to the Eq. (\ref{si}) given in Definition \ref{hse}.}
\begin{proof}
From Eq. (\ref{eq.si}), we can rewrite structural information of $G$ w.r.t all nodes of $\mathcal{T}$ at height $h$ as follows:
\begin{align}
\label{eq.si_at_k}
    \mathcal{H}^{\mathcal{T}}(G;h) &= -\frac{1}{V}\sum_{k=1}^{N_h}g^h_k \log_2 \frac{V_k^h}{V^{h-1}_{k^-}} \nonumber  \\
    &=-\frac{1}{V}[\sum_{k=1}^{N_h}g^h_k \log_2 V_k^h - \sum_{k=1}^{N_h}g^h_k \log_2 V^{h-1}_{k^-}].
\end{align}
Then, the $H$-dimensional structural information of $G$ can be written as
\begin{align}
    \mathcal{H}^{\mathcal{T}}(G) = -\frac{1}{V}[\sum_{h=1}^{H}\sum_{k=1}^{N_h}g^h_k \log_2 V^h_k - \sum_{h=1}^{H}\sum_{k=1}^{N_h}g^h_k \log_2 V^{h-1}_{k^-}].
\end{align}
Since at height $h$,
\begin{align}
    g^h_k &= V^h_k - \sum_{i=1}^N \sum_{j \in \mathcal{N}(i)}\mathbf{I}(i\in T_{\alpha^h_k})\mathbf{I}(j\in T_{\alpha^h_k})w_{ij} \nonumber \\
    &=V^h_k - \sum_{i=1}^N \sum_{j \in \mathcal{N}(i)}S^h_{ik}S^h_{jk}w_{ij},
\end{align}
where we use $S^h_{ik}$ to represent $\mathbf{I}(i\in T_{\alpha^h_k})$, and $T_{\alpha^h_k}$ is the subset of $\mathcal{V}$ corresponds the $k$-th node in height $h$.
\begin{align}
    V^h_k &= \sum_{i=1}^N\mathbf{I}(i\in T_{\alpha^h_k})d_i = \sum_{i=1}^N S^h_{ik} d_i.
\end{align}
We then have the following equation:
\begin{align}
    \sum_{h=1}^H\sum_{k=1}^{N_h}g^h_k \log_2 V^h_k = \sum_{h=1}^H\sum_{k=1}^{N_h}V^h_k \log_2 V^h_k - \sum_{h=1}^H\sum_{k=1}^{N_h}(\log_2 V^h_k) \sum_{i=1}^N \sum_{j \in \mathcal{N}(i)} S^h_{ik}S^h_{jk} w_{ij}.
\end{align}
Similarly, 
\begin{align}
    \sum_{h=1}^H\sum_{k=1}^{N_h}g^h_k \log_2 V^{h-1}_{k^-}= \sum_{h=1}^H\sum_{k=1}^{N_h}V^h_k \log_2 V^{h-1}_{k^-} - \sum_{h=1}^H\sum_{k=1}^{N_h}(\log_2 V^{h-1}_{k^-}) \sum_{i=1}^N \sum_{j \in \mathcal{N}(i)} S^h_{ik}S^h_{jk}w_{ij}.
\end{align}
To find the ancestor of node $k$ in height $h-1$, we utilize the assignment matrix $C^h$ as follows:
\begin{align}
    V^{h-1}_{k^-} = \sum_{k'=1}^{N_{h-1}}\mathbf{C}^h_{k{k'}}V^{h-1}_{k'}.
\end{align}
We can easily verify that
\begin{align}
    S^h_{ik} &= \mathbf{I}(i\in T_{\alpha^h_k}) = \sum_{\alpha^{h-1}_k}^{N_{H-1}}...\sum_{\alpha^{h+1}_k}^{N_{h+1}}\mathbf{I}(\{i\} \subseteq T_{\alpha^{H-1}_k})\mathbf{I}(T_{\alpha^{H-1}_k}\subseteq T_{\alpha^{H-2}_k})...(T_{\alpha^{h+1}_k}\subseteq T_{\alpha^{h}_k})\nonumber \\
    &= \sum_{j_{H-1}}^{N_{H-1}}...\sum_{j_{h+1}}^{N_{h+1}} \mathbf{C}^H_{ij_{H-1}}\mathbf{C}^{H-1}_{j_{H-1}j_{H-2}}...\mathbf{C}^{h+1}_{j_{h+1}k}.
\end{align}
Since $\mathbf{C}^{H+1}=\mathbf{I}_N$, we have $\mathbf{S}^h = \prod_{h=H+1}^{h+1}\mathbf{C}^h$.

Utilizing the equations above, we have completed the proof. 
\end{proof}

\subsection{Proof of Theorem \ref{theorem.approx}}
\label{proof.theoremapprox}

\textbf{Theorem 4.7} (Bound)
\emph{ For any graph $G$, $\mathcal{T}^*$ is the optimal partitioning tree given by  \citet{li2016structural},
 and $\mathcal{T}_{\operatorname{net}}^*$ with height $H$ is the partitioning tree given from Eq. (\ref{eq.opt_z}).
For any pair of leaf embeddings $\boldsymbol z_i$ and $\boldsymbol z_j$ of $\mathcal{T}_{\operatorname{net}}^*$, 
 there exists bounded real functions $\{f_h\}$ and constant $c$,
 such that for any $0 < \epsilon < 1$, $\tau \leq \mathcal{O}(1/ \ln[(1-\epsilon)/\epsilon])$ and  $\frac{1}{1 + \exp\{-(f_h(\boldsymbol z_i, \boldsymbol z_j) - c) / \tau\}} \geq 1 - \epsilon$ satisfying $i\overset{h}{\sim} j$, we have} 
      \vspace{-0.05in}
    \begin{align}
        \lvert \mathcal{H}^{\mathcal{T}^*}(G) - \mathcal{H}^{\mathcal{T}_{\operatorname{net}}^*}(G) \rvert \leq \mathcal{O}(\epsilon).
    \end{align}
We first prove the existence of $\{f_h\}$ by giving a construction process. Fix $0 < \epsilon < 1$, we have
\begin{align}
    &\sigma^h_{ij}=\frac{1}{1 + \exp\{-(f_h(\boldsymbol z_i, \boldsymbol z_j) - c) / \tau\}} \geq 1 - \epsilon \\
    &\Rightarrow f_h(\boldsymbol z_i, \boldsymbol z_j) \geq c + \tau \ln \frac{1-\epsilon}{\epsilon} \leq 1 + c ,
    \label{fc}
\end{align}
since $\tau \leq \mathcal{O}(1/ \ln[(1-\epsilon)/\epsilon])$.
Similarly, when $i\overset{h}{\sim} j$ fails, we have $\sigma^h_{ij}\leq \epsilon \Rightarrow f_h(\boldsymbol z_i, \boldsymbol z_j) \leq c - \tau \ln \frac{1-\epsilon}{\epsilon} \geq c - 1$. Without loss of generality, we let $c=0$ and find that if $i\overset{h}{\sim} j$ holds, $f_h \geq 1$, otherwise, $f_h \leq -1$. If we rewrite $f_h$ as a compose of scalar function $f_h = f(g_h(\boldsymbol z_i, \boldsymbol z_j)) + u(g_h(\boldsymbol z_i, \boldsymbol z_j)): \mathbb{R}\rightarrow \mathbb{R}$, where $g_h$ is a scalar function and $u$ is a step function as $u(x)=
\begin{cases}
1 & x \geq 0\\
-1 & x < 0
\end{cases}$. If $f$ has a discontinuity at $0$ then the construction is easy. At level $h$, we find a node embedding $\boldsymbol a^h$ that is closest to the root $\boldsymbol z_o$ and denote the lowest common ancestor of $\boldsymbol z_i$ and $\boldsymbol z_j$ in a level less than $h$ as $\boldsymbol a^h_{ij}$. Then set $g_h(\boldsymbol z_i, \boldsymbol z_j) = d_\mathbb{L}(\boldsymbol z_o, \boldsymbol a^h_{ij}) - d_\mathbb{L}(\boldsymbol z_o, \boldsymbol a^h)$. Clearly $g_h(\boldsymbol z_i, \boldsymbol z_j)  \geq 0$ if $i\overset{h}{\sim} j$ holds. Then we set $f(\cdot)=\sinh(\cdot)$, i.e. $f_h(\boldsymbol z_i, \boldsymbol z_j) = \sinh(g_h(\boldsymbol z_i, \boldsymbol z_j)) + u(g_h(\boldsymbol z_i, \boldsymbol z_j))$.

If $i\overset{h}{\sim} j$ holds, clearly $f_h \geq 1$, otherwise, $f_h \leq -1$. Since the distance between all pairs of nodes is bounded, $g_h$ is also bounded. All the properties are verified, so such $\{f_h\}$ exist.

Then we have the following lemma.
\begin{lemma}
\label{lemma.kse2}
    For a weighted Graph $G=(V, E)$ with a weight function $w$, and a partitioning tree $\mathcal{T}$ of $G$ with height $H$, we can rewrite the formula of $H$-dimensional structural information of $G$ w.r.t $\mathcal{T}$ as follows:
    \begin{align}
    \label{kse2}
        \mathcal{H}^\mathcal{T}(G)=&-\frac{1}{V} 
            \sum_{h=1}^H\sum_{i=1}^N (d_i-\sum_{j\in \mathcal{N}(i)}\mathbf{I}(i\overset{h}{\sim} j)w_{ij})
            \cdot \log_2(\frac{\sum_l^N\mathbf{I}(i \overset{h}{\sim} l)d_l}{\sum_l^N\mathbf{I}(i \overset{h-1}{\sim} l)d_l}),
    \end{align}
    where $\mathbf{I}(\cdot)$ is the indicator function. 
\end{lemma}
\begin{proof}
    We leave the proof of Lemma \ref{lemma.kse2} in Appendix \ref{proof.kse2}.
\end{proof}

% \begin{lemma}
% \label{lemma:eq}
%     $\mathbf{I}(i \overset{h}{\sim} j)$ can be represent related to the assignment matrix $\mathbf{C}^h$ as
%     \begin{align}
%         \mathbf{I}(i \overset{h}{\sim} j) = \sum_{k=1}^{N_h}s^h_{ik}s^h_{jk},
%     \end{align}
%     where $s^h_{ik}$ is from Equation \ref{eq.s}.
% \end{lemma}
% \begin{proof}
%     At height $h$ of tree $\mathcal{T}$, $\mathbf{I}(i \overset{h}{\sim} j) = \sum_{k=1}^{N_h}\mathbf{I}(i\in T_{\alpha^h_k})\mathbf{I}(j\in T_{\alpha^h_k})=\sum_{k=1}^{N_h}s^h_{ik}s^h_{jk}$, complete the proof.
% \end{proof}

Now we start our proof of Theorem \ref{theorem.approx}.
\begin{proof}
We then focus on the absolute difference between $H^{\mathcal{T}^*}(G) $ and $H^{\mathcal{T}^*_{\operatorname{net}}}(G)$
\begin{align}
\label{eq.ineq}
    \lvert \mathcal{H}^{\mathcal{T}^*}(G) - H^{\mathcal{T}^*_{\operatorname{net}}}(G) \rvert &\leq \lvert \mathcal{H}^{\mathcal{T}^*}(G) - \mathcal{H}^{\mathcal{T}_{\operatorname{net}}}(G; \mathbf{Z}^*; \Theta^*) \rvert + \lvert \mathcal{H}^{\mathcal{T}_{\operatorname{net}}}(G; \mathbf{Z}^*; \Theta^*) - H^{\mathcal{T}^*_{\operatorname{net}}}(G) \rvert \nonumber \\
    &\leq 2 \sup_{\mathbf{Z}, \mathbf{\Theta}}\lvert \mathcal{H}^{\mathcal{T}}(G) - \mathcal{H}^{\mathcal{T}_{\operatorname{net}}}(G; \mathbf{Z}; \Theta) \rvert,
\end{align}
where $\mathcal{T}=\operatorname{decode}(\mathbf{Z})$.
Recall the Eq. (\ref{kse2}), we can divide them into four corresponding parts, denoted as $A$, $B$, $C$, and $D$ respectively. Then, we have
\begin{align}
    \lvert \mathcal{H}^{\mathcal{T}}(G) - \mathcal{H}^{\mathcal{T}_{\operatorname{net}}}(G; \mathbf{Z}; \Theta) \rvert &= -\frac{1}{\operatorname{Vol}(G)}[A + B + C + D] \nonumber \\
    &\leq -\frac{1}{\operatorname{Vol}(G)} [\lvert A \rvert + \lvert B \rvert + \lvert C \rvert + \lvert D \rvert].
\end{align}
For part $A$, give fixed $i$ and $k$, we assume $i \overset{h}{\sim} l$ holds for some $l_{1},...,l_{m_i}$.
\begin{align}
    \lvert A \rvert &= \lvert \sum_{h=1}^H\sum_i^N d_i \log_2(\sum_l^N \mathbf{I}(i \overset{h}{\sim} l) d_l)-\sum_{h=1}^H\sum_i^N d_i \log_2(\sum_l^N \sigma_{il}^h d_l) \rvert \nonumber \\
    &\leq \lvert \sum_{h=1}^H\sum_i^N d_i\log_2 \frac{\sum_{c=1}^{m_i} d_{l_c}}{\sum_{c=1}^{m_i} (1-\epsilon)d_{l_c} + \sum_{\overset{-}{c}}\epsilon d_{l_{\overset{-}{c}}}} \rvert \nonumber \\
    &\leq \lvert \sum_{h=1}^H\sum_i^N d_i\log_2 \frac{\sum_{c=1}^{m_i} d_{l_c}}{\sum_{c=1}^{m_i}(1-\epsilon)d_{l_c}} \rvert \nonumber \\
    &=\lvert \sum_{h=1}^H\sum_i^N d_i \log_2 \frac{1}{1-\epsilon} \rvert \nonumber \\
    &= H\operatorname{Vol}(G)\log_2 \frac{1}{1-\epsilon}.
\end{align}
For part B, we still follow the assumption above and assume $i \overset{h}{\sim} j$ holds for some $j_e,...,j_{n_i}$ in the neighborhood of $i$.
\begin{align}
    \lvert B \rvert &= \lvert-\sum_{h=1}^H\sum_i^N[\log_2(\sum_l^N\mathbf{I}(i\overset{h}{\sim} l)d_l)\cdot \sum_{j\in \mathcal{N}(i)}\mathbf{I}(i\overset{h}{\sim} j)w_{ij}] +\sum_{h=1}^H\sum_i^N[\log_2(\sum_l^N\sigma^h_{il}d_l)\cdot \sum_{j\in \mathcal{N}(i)}\sigma^h_{ij}w_{ij} ] \rvert \nonumber \\
    &\leq \lvert\sum_{h=1}^H\sum_i^N\log_2(\sum_{c=1}^{m_i} d_{l_c})\cdot \sum_{e=1}^{n_i}w_{ij_e} -\log_2(\sum_{c=1}^{m_i}(1-\epsilon)d_{l_c}+\sum_{\overset{-}{c}}\epsilon d_{l_{\overset{-}{c}}})\cdot [\sum_{e=1}^{n_i}(1-\epsilon)w_{ij_e} + \sum_{\overset{-}{e}}\epsilon w_{ij_{\overset{-}{e}}}] \rvert \nonumber \\
    & \leq \lvert\sum_{h=1}^H\sum_i^N\log_2(\sum_{c=1}^{m_i} d_{l_c})\cdot \sum_{e=1}^{n_i}w_{ij_e} -\log_2(\sum_{c=1}^{m_i}(1-\epsilon)d_{l_c})\cdot \sum_{e=1}^{n_i}(1-\epsilon)w_{ij_e} \rvert \nonumber \\
    &=\lvert \sum_{h=1}^H\sum_i^N (\sum_{e=1}^{n_i}w_{ij_e})[\log_2(\sum_{c=1}^{m_i} d_{l_c} - (1-\epsilon)\log_2(1-\epsilon) - (1-\epsilon)\log_2(\sum_{c=1}^{m_i} d_{l_c}]  \rvert 
    \nonumber \\
    &= \lvert \sum_{h=1}^H\sum_i^N (\sum_{e=1}^{n_i}w_{ij_e})[\epsilon \log_2(\sum_{c=1}^{m_i} d_{l_c} - (1-\epsilon)\log_2(1-\epsilon)]  \rvert 
    \nonumber \\
    &\leq (1-\epsilon)\log_2 \frac{1}{1-\epsilon}\sum_{h=1}^H\sum_i^N\sum_{e=1}^{n_i}w_{ij_e}\nonumber \\
    &\leq (1-\epsilon)\log_2 \frac{1}{1-\epsilon}\sum_{h=1}^H\sum_i^N d_i\nonumber \\
    &=(1-\epsilon)H V\log_2 \frac{1}{1-\epsilon} .
\end{align}
Similarly, we can get the same results for parts $C$ and $D$ as $A$ and $B$ respectively. Then,
\begin{align}
    \lvert \mathcal{H}^{\mathcal{T}}(G) - \mathcal{H}^{\mathcal{T}_{\operatorname{net}}}(G; \mathbf{Z}; \Theta) \rvert &\leq 2H(2-\epsilon)\log_2\frac{1}{1-\epsilon}.
\end{align}
Substituting into Equation.\ref{eq.ineq}, we obtain
\begin{align}
    \lvert \mathcal{H}^{\mathcal{T}^*}(G) - H^{\mathcal{T}^*_{\operatorname{net}}}(G) \rvert &\leq 4H(2-\epsilon)\log_2\frac{1}{1-\epsilon}.
\end{align}
Since the limitation is a constant number,
\begin{align}
    \lim_{\epsilon \rightarrow 0} \frac{(2-\epsilon)\log_2\frac{1}{1-\epsilon}}{\epsilon}=2,
\end{align}
we have $\lvert H^{\mathcal{T}^*}(G) - H^{\mathcal{T}^*_{\operatorname{net}}}(G) \rvert \leq \mathcal{O}(\epsilon)$, which completes the proof. $\hfill\square$
\end{proof}

\subsection{Proof of Lemma \ref{lemma.add}}
\label{proof.add}
\textbf{Lemma 4.5} (Additive)
\emph{
    The one-dimensional structural entropy of $G$ can be decomposed as follows
    \begin{align}
        \mathcal{H}^1(G) = \sum_{h=1}^H\sum_{j=1}^{N_{h-1}}\frac{V^{h-1}_j}{V}E([\frac{C^h_{kj}V^h_k}{V^{h-1}_j}]_{k=1,...,N_h}),
    \end{align}
    where 
      $  E(p_1, ..., p_n) = -\sum_{i=1}^n p_i\log_2 p_i$
 is the entropy. 
}

\begin{proof}
To verify the equivalence to the one-dimensional structural entropy, we expand the above formula.
\begin{align}
    \mathcal{H}(G)&=-\sum_{h=1}^H\sum_{j=1}^{N_{h-1}}\frac{V^{h-1}_j}{V}\sum_{k=1}^{N_h}\frac{C^h_{kj}V^h_k}{V^{h-1}_j}\log_2 \frac{C^h_{kj}V^h_k}{V^{h-1}_j} \\
    &= -\frac{1}{V}\sum_{h=1}^H \sum_{k=1}^{N_h}\sum_{j=1}^{N_{h-1}} C^h_{kj}V^h_k \log_2 \frac{C^h_{kj}V^h_k}{V^{h-1}_j}\\
    &=-\frac{1}{V}\sum_{h=1}^H \sum_{k=1}^{N_h}V^h_k \log_2\frac{V^h_k}{V^{h-1}_{k^-}}.
\end{align}
We omit some $j$ that make $C^h_{kj}=0$ so that the term $V^{h-1}_j$ only exists in the terms where $C^h_{kj}=1$, meaning that $V^{h-1}_j$ exists if and only if $V^{h-1}_j=V^{h-1}_{k^-}$ since we summing over all $k$. Continue the process that
\begin{align}
    \mathcal{H}(G) &= -\frac{1}{V}\sum_{h=1}^H \sum_{k=1}^{N_h}V^h_k \log_2\frac{V^h_k}{V} + \frac{1}{V}\sum_{h=1}^H \sum_{k=1}^{N_h}V^h_k \log_2\frac{V^{h-1}_{k^-}}{V} \\
    &=-\frac{1}{V}\sum_{i=1}^N d_i\log_2\frac{d_i}{V} -\frac{1}{V}\sum_{h=1}^{H-1} \sum_{k=1}^{N_h}V^h_k \log_2\frac{V^h_k}{V} + \frac{1}{V}\sum_{h=1}^H \sum_{k=1}^{N_h}V^h_k \log_2\frac{V^{h-1}_{k^-}}{V} \\
    &=-\frac{1}{V}\sum_{i=1}^N d_i\log_2\frac{d_i}{V} -\frac{1}{V}\sum_{h=1}^{H-1} \sum_{k=1}^{N_h}V^h_k \log_2\frac{V^h_k}{V} + \frac{1}{V}\sum_{h=2}^H \sum_{k=1}^{N_h}V^h_k \log_2\frac{V^{h-1}_{k^-}}{V}.
\end{align}
From the first line to the second line, we separate from the term when $h=H$. For the third line, we eliminate the summation term in $h=1$, since when $h=1$, in the last term, $\log_2\frac{V^{0}_{k^-}}{V}=\log_2\frac{V}{V}=0$.

Then let us respectively denote $-\sum_{h=1}^{H-1} \sum_{k=1}^{N_h}V^h_k \log_2\frac{V^h_k}{V}$ and $\sum_{h=2}^H \sum_{k=1}^{N_h}V^h_k \log_2\frac{V^{h-1}_{k^-}}{V}$ as $A$ and $B$, using the trick about $C^h_{kj}$ like above, we have
\begin{align}
    A+B &= \sum_{h=2}^{H} [(\sum_{k=1}^{N_h}V^h_k \log_2\frac{V^{h-1}_{k^-}}{V}) - (\sum_{j=1}^{N_{h-1}}V^{h-1}_j \log_2\frac{V^{h-1}_j}{V})] \\
    &= \sum_{h=2}^{H}[ \sum_{k=1}^{N_h}V^h_k \log_2\frac{V^{h-1}_{k^-}}{V} - \sum_{j=1}^{N_{h-1}}\sum_{k=1}^{N_h}C^h_{kj}V^h_k \log_2\frac{V^{h-1}_j}{V})] \\
    &=\sum_{h=2}^{H} [\sum_{k=1}^{N_h}V^h_k \log_2\frac{V^{h-1}_{k^-}}{V} - \sum_{k=1}^{N_h}V^h_k \log_2\frac{V^{h-1}_{k^-}}{V}] \\
    &=0 .\nonumber
\end{align}
Thus 
\begin{align}
    \mathcal{H}(G)=-\frac{1}{V}\sum_{i=1}^N d_i\log_2\frac{d_i}{V},
\end{align}
which is exactly the one-dimensional structural entropy $\mathcal{H}^1(G)$ of $G$. The proof is completed.
\end{proof}

\subsection{Proof of Theorem \ref{theorem.conductance}}
\label{proof.conductance}
\textbf{Theorem 4.6} (Connection to Graph Clustering)
\emph{Given a graph $G=(\mathcal{V}, \mathcal{E})$ with  $w$, the normalized $H$-structural entropy of graph $G$ is defined as
    $\tau(G;H) = {\mathcal{H}^H(G)}/{\mathcal{H}^1(G)}$, and 
$\Phi(G)$ is the graph conductance. 
    With the additivity of DSE \emph{(Lemma \ref{lemma.add})}, the following inequality holds,
    %Then the normalized $H$-dimensional structural information of $G$ w.r.t. a partitioning tree $\mathcal{T}$ satisfies the following property:
    \begin{align}
        \tau(G; H) \geq \Phi(G).
    \end{align}
    }
\begin{proof}
Recall the formula of $H$-dimensional structural information of $G$ for a partitioning tree $\mathcal{T}$ in Definition \ref{DSE}. For the conductance $\phi_{h,k}$ of $k$-th node in $\mathcal{T}$ as height $h$, following the definition of graph conductance, we have
\begin{align}
    \phi_{h,k} = \frac{\sum_{i\in T_{h,k}, j \notin T_{h,k}}w_{ij}}{\min\{ V^h_k, V-V^h_k \}} 
    = \frac{V^h_k-\sum_{(i,j)\in\mathcal{E}}s^h_{ik}s^h_{jk}w_{ij}}{V^h_k}.
\end{align}
Without loss of generality, we assume that $V^h_k \leq \frac{1}{2}V$ such that $\min\{ V^h_k, V-V^h_k \}=V^h_k$.
From Definition \ref{DSE}, we have
\begin{align}
    \mathcal{H}^\mathcal{T}(G) &= -\frac{1}{V}\sum_{h=1}^H\sum_{k=1}^{N_h} \phi_{h,k}V^h_k\log_2\frac{V^h_k}{V^{h-1}_{k^-}} \\
    &\geq -\Phi(G)\sum_{h=1}^H \sum_{k=1}^{N_h}\frac{V^h_k}{V}\log_2\frac{V^h_k}{V^{h-1}_{k^-}} \\
    &=-\Phi(G)\sum_{h=1}^H \sum_{k=1}^{N_h}\frac{\sum_{j=1}^{N_{h-1}}C^h_{kj}V^h_k}{V}\log_2\frac{V^h_k}{V^{h-1}_{k^-}} \\
    &=-\Phi(G)\sum_{h=1}^H \sum_{k=1}^{N_h} \sum_{j=1}^{N_{h-1}} \frac{V^{h-1}_j}{V}\frac{C^h_{kj}V^h_k}{V^{h-1}_j}\log_2\frac{V^h_k}{V^{h-1}_{k^-}} \\
    &=-\Phi(G)\sum_{h=1}^H \sum_{j=1}^{N_{h-1}} \frac{V^{h-1}_j}{V} \sum_{k=1}^{N_h} \frac{C^h_{kj}V^h_k}{V^{h-1}_j}\log_2\frac{V^h_k}{\sum_{m=1}^{N_{h-1}}C^h_{km}V^{h-1}_{m}} \\
    &=-\Phi(G)\sum_{h=1}^H \sum_{j=1}^{N_{h-1}} \frac{V^{h-1}_j}{V} \sum_{k=1}^{N_h} \frac{C^h_{kj}V^h_k}{V^{h-1}_j}\log_2\frac{C^h_{kj}V^h_k}{C^h_{kj}\sum_{m=1}^{N_{h-1}}C^h_{km}V^{h-1}_{m}}
\end{align}

\begin{align}
 &-\Phi(G)\sum_{h=1}^H \sum_{j=1}^{N_{h-1}} \frac{V^{h-1}_j}{V} \sum_{k=1}^{N_h} \frac{C^h_{kj}V^h_k}{V^{h-1}_j}\log_2\frac{C^h_{kj}V^h_k}{C^h_{kj}\sum_{m=1}^{N_{h-1}}C^h_{km}V^{h-1}_{m}} \\
    =&-\Phi(G)\sum_{h=1}^H \sum_{j=1}^{N_{h-1}} \frac{V^{h-1}_j}{V} \sum_{k=1}^{N_h} \frac{C^h_{kj}V^h_k}{V^{h-1}_j}\log_2\frac{C^h_{kj}V^h_k}{V^{h-1}_j} \\
    =&\Phi(G)\sum_{h=1}^H\sum_{j=1}^{N_{h-1}}\frac{V^{h-1}_j}{V}\operatorname{Ent}([\frac{C^h_{kj}V^h_k}{V^{h-1}_j}]_{k=1,...,N_h}) \\
    =&\Phi(G)\mathcal{H}^1(G).
\end{align}
To get the final results, we use the inverse trick about $C^h_{kj}$ mentioned in Appendix \ref{proof.add}.
So the normalized $H$-dimensional structural information of $G$ w.r.t. $\mathcal{T}$ satisfies
\begin{align}
    \tau(G;\mathcal{T}) = \frac{\mathcal{H}^\mathcal{T}(G)}{\mathcal{H}^1(G)} \geq \Phi(G).
\end{align}
Since this inequality holds for every $H$-height partitioning tree $\mathcal{T}$ of $G$, $\tau(G;H)\geq \Phi(G)$ holds. 
\end{proof}

\subsection{Proof of Lemma \ref{lemma.kse2}}
\label{proof.kse2}
\textbf{Lemma A.1}
    \emph{For a weighted Graph $G=(V, E)$ with a weight function $w$, and a partitioning tree $\mathcal{T}$ of $G$ with height $H$, we can rewrite the formula of $H$-dimensional structural information of $G$ w.r.t $\mathcal{T}$ as follows:
    \begin{align}
        \mathcal{H}^\mathcal{T}(G;H)=&-\frac{1}{V} 
            \sum_{h=1}^H\sum_{i=1}^N (d_i-\sum_{j\in \mathcal{N}(i)}\mathbf{I}(i\overset{h}{\sim} j)w_{ij})
            \cdot \log_2(\frac{\sum_l^N\mathbf{I}(i \overset{h}{\sim} l)d_l}{\sum_l^N\mathbf{I}(i \overset{h-1}{\sim} l)d_l}),
    \end{align}
    where $\mathbf{I}(\cdot)$ is the indicator function.}
\begin{proof}
From Equation.\ref{eq.si}, we can rewrite structural information of $G$ w.r.t all nodes of $\mathcal{T}$ at height $h$ as follows:
\begin{align}
\label{eq.si_at_k}
    H^{\mathcal{T}}(G;h) &= -\frac{1}{V}\sum_{k=1}^{N_h}g^h_k \log_2 \frac{V^h_k}{V^{h-1}_{k^-}} \nonumber  \\
    &=-\frac{1}{V}[\sum_{k=1}^{N_h}g^h_k \log_2 V^h_k - \sum_{k=1}^{N_h}g^h_k \log_2 V^{h-1}_{k^-}].
\end{align}
Then the $H$-dimensional structural information of $G$ can be written as
\begin{align}
    \mathcal{H}^{\mathcal{T}}(G) = -\frac{1}{V}[\sum_{h=1}^H\sum_{k=1}^{N_h}g^h_k \log_2 V^h_k - \sum_{h=1}^H\sum_{k=1}^{N_h}g^h_k \log_2 V^{h-1}_{k^-}].
\end{align}

Since
\begin{align}
    g^h_k &= V^h_k - \sum_{i=1}^N \sum_{j \in \mathcal{N}(i)}\mathbf{I}(i\in T_{\alpha^h_k})\mathbf{I}(j\in T_{\alpha^h_k})w_{ij},
\end{align}
\begin{align}
    V^h_k = \sum_i^N\mathbf{I}(i\in T_{\alpha^h_k})d_i,
\end{align}
we have the following equation with the help of the equivalence relationship $i\overset{h}{\sim} j$:
\begin{align}
    \sum_{h=1}^H\sum_{k=1}^{N_h}g^h_k \log_2 V^h_k = \sum_{h=1}^H\sum_{k=1}^{N_h}V^h_k \log_2 V^h_k - \sum_{h=1}^H\sum_{k=1}^{N_h}(\log_2 V^h_k) \sum_{i=1}^N \sum_{j \in \mathcal{N}(i)} \mathbf{I}(i\overset{h}{\sim} j)w_{ij}.
\end{align}
Then we focus on the above equation term by term. If we fix the height of $h$ and the node $i$ of $G$, then the node $\alpha$ to which $i$ belongs and its ancestors $\alpha^-$ in $\mathcal{T}$ are also fixed. Let's denote $V^h_k$ as $V_\alpha$ where $h(\alpha)=h$.
\begin{align}
    \sum_{h=1}^H\sum_{h(\alpha)=h}V_\alpha \log_2 V_\alpha &=  \sum_{h=1}^H\sum_{h(\alpha)=h}\sum_{i=1}^N\mathbf{I}(i\in T_\alpha)d_i \log_2(\sum_{l=1}^N\mathbf{I}(l\in T_\alpha)d_l) \nonumber \\
    &= \sum_{h=1}^H\sum_{i=1}^Nd_i \log_2 (\sum_{l=1}^N\mathbf{I}(i\overset{h}{\sim} l)d_l),
\end{align}
\begin{align}
    \sum_{h=1}^H\sum_{h(\alpha)=h}(\log_2 V_\alpha) \sum_{i=1}^N \sum_{j \in \mathcal{N}(i)} \mathbf{I}(i\overset{h}{\sim} j)w_{ij} &= \sum_{h=1}^H\sum_{i=1}^N [\log_2 (\sum_{l=1}^N\mathbf{I}(i\overset{h}{\sim} l)d_l)] \cdot \sum_{j \in \mathcal{N}(i)}\mathbf{I}(i\overset{h}{\sim} j)w_{ij}.
\end{align}
Similarly, 
\begin{align}
    \sum_{h=1}^H\sum_{h(\alpha)=h}g_\alpha \log_2 V_{\alpha^-}= \sum_{h=1}^H\sum_{h(\alpha)=h}V_\alpha \log_2 V_{\alpha^-} - \sum_{h=1}^H\sum_{h(\alpha)=h}(\log_2 V_{\alpha^-}) \sum_i^N \sum_{j \in \mathcal{N}(i)} \mathbf{I}(i\overset{h}{\sim} j)w_{ij},
\end{align}

\begin{align}
    \sum_{h=1}^H\sum_{h(\alpha)=h}V_\alpha \log_2 V_\alpha &=  \sum_{h=1}^H\sum_{h(\alpha)=h}\sum_{i=1}^N\mathbf{I}(i\in T_\alpha)d_i \log_2(\sum_{l=1}^N\mathbf{I}(l\in T_{\alpha^-})d_l) \nonumber \\
    &= \sum_{h=1}^H\sum_{i=1}^Nd_i \log_2 (\sum_{l=1}^N\mathbf{I}(i\overset{h-1}{\sim} l)d_l),
\end{align}

\begin{align}
    \sum_{h=1}^H\sum_{h(\alpha)=h}(\log_2 V_{\alpha^-}) \sum_{i=1}^N \sum_{j \in \mathcal{N}(i)} \mathbf{I}(i\overset{h}{\sim} j)w_{ij} = \sum_{h=1}^H\sum_{i=1}^N [\log_2 (\sum_{l=1}^N\mathbf{I}(i\overset{h-1}{\sim} l)d_l)] \cdot \sum_{j \in \mathcal{N}(i)}\mathbf{I}(i\overset{h}{\sim} j)w_{ij}.
\end{align}
Utilizing the equations above, we have completed the proof.
\end{proof}

\subsection{Proof of Theorem \ref{theorem.flex}}
\label{proof.flex}
\textbf{Theorem 4.8} (Flexible)
    \emph{$\forall G$, given a partitioning tree $\mathcal{T}$ and adding a node $\beta$ to get a relaxed partitioning tree $\mathcal{T}'$ , the structural information remains unchanged, $H^{\mathcal{T}}(G)=H^{\mathcal{T}'}(G)$, if  one of the following conditions holds:
    \begin{enumerate}
      \item $\beta$ as a leaf node, and the corresponding node subset $T_\beta$ is an empty set.
      \item $\beta$ is inserted between node $\alpha$ and its children nodes so that the corresponding node subsets $T_\beta = T_\alpha$.
\end{enumerate}}

\begin{proof}
    For the first case, since $\beta$ is a leaf node corresponds an empty subset of $\mathcal{V}$, $g_\beta=0$, then the structural information of $\beta$ is $\mathcal{H}^{\mathcal{T}}(G;\beta)=0$.
    For the second case, we notice the term $\log_2\frac{V_\alpha}{V_{\alpha^-}}$ in Equation \ref{eq.si}. Since the volume of $T_\alpha$ equals the one of $T_\beta$, the term vanishes to $0$.
\end{proof}

\section{Hyperbolic Space}

\subsection{Riemannian Manifold}
Riemannian manifold is a real and smooth manifold $\mathbb{M}$ equipped with Riemannian metric tensor $g_{\boldsymbol x}$ on the tangent space $\mathcal{T}_{\boldsymbol{x}}\mathbb{M}$ at point $\boldsymbol{x}$.
A Riemannian metric assigns to each $\boldsymbol{x} \in \mathbb{M}$ a positive-definite inner product $g_{\boldsymbol{x}}: \mathcal{T}_{\boldsymbol{x}}\mathbb{M} \times \mathcal{T}_{\boldsymbol{x}}\mathbb{M} \rightarrow \mathbb{R}$, which induces a norm defined as $\lvert \cdot \rvert: \boldsymbol{v} \in \mathcal{T}_{\boldsymbol{x}}\mathbb{M} \mapsto \sqrt{g_{\boldsymbol{x}}(\boldsymbol{v}, \boldsymbol{v})} \in \mathbb{R}$.

An exponential map at point $\boldsymbol x \in \mathbb{M}$ is denoted as $\operatorname{Exp}_{\boldsymbol x}(\cdot): \boldsymbol u \in \mathcal{T}_{\boldsymbol{x}}\mathbb{M} \mapsto \operatorname{Exp}_{\boldsymbol x}(\boldsymbol u) \in \mathbb{M}$. It takes a tangent vector $\boldsymbol u$ in the tangent space at $\boldsymbol x$ and transforms $\boldsymbol x$ along the geodesic starting at $\boldsymbol x$ in the direction of $\boldsymbol u$.

A logarithmic map at point $\boldsymbol x$ is the inverse of the exponential map $\boldsymbol x$, which maps the point $\boldsymbol y \in \mathbb{M}$ to a vector $\boldsymbol{v}$ in the tangent space at $\boldsymbol x$.
\subsection{Models of Hyperbolic Space}
\paragraph{Poincar\'{e} ball model} The $d$-dimensional Poincar\'{e} ball model with constant negative curvature $\kappa$, is defined within a $d$-dimensional hypersphere, formally denoted as $\mathbb{B}^d_\kappa = \{\boldsymbol{x} \in \mathbb{R}^d \lvert \lVert \boldsymbol{x} \rVert^2=-\frac{1}{\kappa}\}$. The Riemannian metric tensor at $\boldsymbol{x}$ is 
$g_{\boldsymbol x}^\kappa=(\lambda_{\boldsymbol{x}}^\kappa)^2g_\mathbb{E}=\frac{4}{(1+\kappa\lVert \boldsymbol{x} \rVert^2)^2}g_\mathbb{E}$, 
where $g_\mathbb{E}=\mathbf{I}$ is the Euclidean metric. The distance function is given by
\begin{align}
     d_\mathbb{B}^\kappa(\boldsymbol{x}, \boldsymbol{y})=\frac{2}{\sqrt{-\kappa}}\tanh^{-1}(\lVert (-\boldsymbol{x}) \oplus_\kappa \boldsymbol{y} \rVert),
\end{align}
where $\oplus_\kappa$ is the M{\"o}bius addition
\begin{align}
    \boldsymbol{x} \oplus_\kappa \boldsymbol{y} = 
    \frac{(1-2\kappa\boldsymbol{x}^\top\boldsymbol{y}-\kappa\lVert \boldsymbol{y} \rVert^2)\boldsymbol{x}+(1+\kappa\lVert \boldsymbol{x} \rVert^2)\boldsymbol{y}}
    {1-2\kappa\boldsymbol{x}^\top\boldsymbol{y}+\kappa^2\lVert \boldsymbol{x} \rVert^2\lVert \boldsymbol{y} \rVert^2}.
\end{align}
The exponential and logarithmic maps of Pincar\'{e} ball model are defined as 
\begin{align}
    \operatorname{Exp}_{\boldsymbol{x}}^\kappa(\boldsymbol{v})&=\boldsymbol{x} \oplus_\kappa(\frac{1}{\sqrt{-\kappa}}\tanh(\sqrt{-\kappa}\frac{\lambda^\kappa_{\boldsymbol{x}}\lVert \boldsymbol{v} \rVert}{2})\frac{\boldsymbol{v}}{\lVert \boldsymbol{v} \rVert}) \\
    \log_{\boldsymbol{x}}^\kappa(\boldsymbol{y})&=\frac{2}{\sqrt{-\kappa}\lambda^\kappa_{\boldsymbol{x}}}
    \tanh^{-1}(\sqrt{-\kappa}\lVert (-\boldsymbol{x}) \oplus_\kappa \boldsymbol{y} \rVert)\frac{(-\boldsymbol{x}) \oplus_\kappa \boldsymbol{y}}{\lVert (-\boldsymbol{x}) \oplus_\kappa \boldsymbol{y} \rVert}.
\end{align}
\paragraph{Lorentz model} The $d$-dimensional Lorentz model equipped with constant curvature $\kappa < 0$ and Riemannian metric tensor $\mathbf{R} =\operatorname{Diag(-1,1,...,1)}$, is defined in $(d+1)$-dimensional Minkowski space whose origin is $(\sqrt{-1/\kappa}, 0,...,0)$, i.e., $\mathbb{L}^d_\kappa=\{\boldsymbol{x}\in \mathbb{R}^{d+1} \lvert \langle \boldsymbol{x}, \boldsymbol{x} \rangle_\mathbb{L}=\frac{1}{\kappa}\}$, where the inner product is $\langle \boldsymbol{x}, \boldsymbol{y} \rangle_\mathbb{L}=-x_0y_0+\sum_{i=1}^dx_iy_i=\boldsymbol{x}^T\mathbf{R}\boldsymbol{y}$. The distance between two points is given by
\begin{align}
    d_\mathbb{L}(\boldsymbol{x}, \boldsymbol{y}) = \operatorname{cosh}^{-1}(-\langle \boldsymbol{x}, \boldsymbol{y} \rangle_\mathbb{L}).
\end{align}
The tangent space at $\boldsymbol{x} \in \mathbb{L}^d_\kappa$ is the set of $\boldsymbol{y} \in \mathbb{L}^d_\kappa$ such that orthogonal to $\boldsymbol{x}$ w.r.t. Lorentzian inner product, denotes as $\mathcal{T}_{\boldsymbol{x}} \mathbb{L}^d_\kappa=\{ \boldsymbol{y} \in \mathbb{R}^{d+1} \lvert \langle \boldsymbol{y}, \boldsymbol{x}\rangle_\mathbb{L}=0 \}$.
The exponential and logarithmic maps at $\boldsymbol x$ are defined as 
\begin{align}
    \operatorname{Exp}_{\boldsymbol x}^\kappa(\boldsymbol u) &= \cosh(\sqrt{-\kappa}\lVert \boldsymbol u\rVert_\mathbb{L})\boldsymbol x + \sinh(\sqrt{-\kappa}\lVert \boldsymbol u\rVert_\mathbb{L})\frac{\boldsymbol u}{\sqrt{-\kappa}\lVert \boldsymbol u\rVert_\mathbb{L}}   \label{expmap} \\
    \log_{\boldsymbol{x}}^\kappa(\boldsymbol{y})&=\frac{\cosh^{-1}(\kappa\langle \boldsymbol x, \boldsymbol y
    \rangle_\mathbb{L})}{\sqrt{(\kappa\langle \boldsymbol x, \boldsymbol y
    \rangle_\mathbb{L})^2-1}}(\boldsymbol y - \kappa\langle \boldsymbol x, \boldsymbol y
    \rangle_\mathbb{L}\boldsymbol x).
\end{align}
\subsection{Tree and Hyperbolic Space}
\label{treehyp}
We denote the embedding of node $i$ of graph $G$ in $\mathbb{H}$ is $\phi_i$. The following definitions and theorem are from \cite{sarkar2012low}.
\begin{definition}[Delaunay Graph]
    Given a set of vertices in $\mathbb{H}$ their Delaunay graph is one where a pair of vertices are neighbors if their Voronoi cells intersect.
\end{definition}

\begin{definition}[Delaunay Embedding of Graphs]
    Given a graph $G$, its Delaunay embedding in $\mathbb{H}$ is an embedding of the vertices such that their Delaunay graph is $G$.
\end{definition}

\begin{definition}[$\beta$ separated cones]
    Suppose cones $C(\overset{\longrightarrow}{\phi_j\phi_i}, \alpha)$ and $C(\overset{\longrightarrow}{\phi_j\phi_x}, \gamma)$ are adjacent with the same root $\varphi j$. Then the cones are $\beta$ separated if the two cones are an angle $2\beta$ apart. That is, for arbitrary points $p \in C(\overset{\longrightarrow}{\phi_j\phi_i}, \alpha)$ and $q \in C (\overset{\longrightarrow}{\phi_j\phi_x}, \gamma)$, the angle $\angle p\phi_jq > 2\beta$.
\end{definition}

\begin{lemma}
    If cones $C(\overset{\longrightarrow}{\phi_j\phi_i}, \alpha)$ and $C (\overset{\longrightarrow}{\phi_j\phi_x}, \gamma)$ are $\beta$ separated and $\phi_r \in C(\overset{\longrightarrow}{\phi_j\phi_i}, \alpha)$ and $\phi_s \in C (\overset{\longrightarrow}{\phi_j\phi_x}, \gamma)$ then there is a constant $\nu$ depending only on $\beta$ such that $|\phi_r\phi_j |_{\mathbb{H}} + |\phi_s\phi_j|_{\mathbb{H}} > |\phi_r\phi_s|_\mathbb{H} > |\phi_r\phi_j |_\mathbb{H} + |\phi_s\phi_j |_{\mathbb{H}} - \nu$.
\end{lemma}

\begin{theorem}[\cite{sarkar2012low}]
    If all edges of $\mathcal{T}$ are scaled by a constant factor $\tau \geq \eta_{\max}$ such that each edge is longer than $\nu\frac{(1+\epsilon)}{\epsilon}$ and the Delaunay embedding of $\mathcal{T}$ is $\beta$ separated, then the distortion over all vertex pairs is bounded by $1+\epsilon$.
\end{theorem}
Following the above Theorem, we know that a tree can be embedded into hyperbolic space with an arbitrarily low distortion.

\subsection{Midpoint in the Manifold}
\label{hyperbolic.midpoint}

Let $(\mathbb{M}, g)$ be a Riemannian manifold, and $\boldsymbol{x}_1, \boldsymbol{x}_2, ..., \boldsymbol{x}_n$ are points on the manifold. The Fr\'{e}chet variance at point $\boldsymbol{\mu}\in \mathbb{M}$ of these points are given by
\begin{align}
    \Psi(\boldsymbol{\mu}) = \sum_i d^2(\boldsymbol{\mu}, \boldsymbol{x}_i).
\end{align}
%If for some $\boldsymbol{\mu}$s locally minimize the variance, they are called Karcher means. 
If there is a point $\boldsymbol{p}$ locally minimizes the variance, it is called Fr\'{e}chet mean. Generally, if we assign each $\boldsymbol{x}_i$ a weight $w_i$, the Fr\'{e}chet mean can be formulated as
\begin{align}
    \boldsymbol{p} = \mathop{\arg\min}_{\boldsymbol{\mu}\in \mathbb{M}} \sum_i w_i d^2(\boldsymbol{\mu}, \boldsymbol{x}_i).
\end{align}
Note that, $d$ is the canonical distance in the manifold. 

\textbf{Theorem 5.2} (Arithmetic Mean as Geometric Centroid) 
\emph{In hyperbolic space $\mathbb L^{\kappa, d_{\mathcal T}}$,
for any set of $\{\boldsymbol z^{h}_i\}$,
the arithmetic mean of
$\boldsymbol z^{h-1}_j =
\frac{1}{\sqrt{-\kappa}} \sum_{i=1}^n \frac{ c_{ij}}{\lvert \lVert \sum_{l=1}^n c_{lj} \boldsymbol z^{h}_l  \rVert_\mathbb{L} \rvert} \boldsymbol z^{h}_i $
is the manifold $\boldsymbol z^{h-1}_j \in \mathbb L^{\kappa, d_{\mathcal T}}$,
and is the close-form solution of the geometric centroid specified in the minimization of Eq. (\ref{centroid}).}
\begin{proof}
Since $d^2_\mathbb{L}(\boldsymbol z^h_i, \boldsymbol z^{h-1}_j) = \lVert \boldsymbol z^h_i- \boldsymbol z^{h-1}_j\rVert^2_\mathbb{L}=\lVert \boldsymbol z^h_i\rVert^2_\mathbb{L} + \lVert \boldsymbol z^{h-1}_j\rVert^2_\mathbb{L} - 2\langle \boldsymbol z^h_i, \boldsymbol z^{h-1}_j 
 \rangle_\mathbb{L} = 2/\kappa - 2\langle \boldsymbol z^h_i, \boldsymbol z^{h-1}_j 
 \rangle_\mathbb{L}$, minimizing Eq. (\ref{centroid}) is equivalent to maximizing $\sum_i c_{ij} \langle \boldsymbol z^h_i, \boldsymbol z^{h-1}_j 
 \rangle_\mathbb{L} =  \langle \sum_i c_{ij} \boldsymbol z^h_i, \boldsymbol z^{h-1}_j \rangle_\mathbb{L}$. To maximize the Lorentz inner product, $\boldsymbol z^{h-1}_j$ must be $\eta \sum_i c_{ij} \boldsymbol z^h_i$ for some positive constant $\eta$. Then the inner product will be
 \begin{align}
     \langle \boldsymbol z^{h-1}_j, \boldsymbol z^{h-1}_j \rangle_\mathbb{L} = \eta^2 \langle \sum_i c_{ij} \boldsymbol z^h_i, \sum_i c_{ij} \boldsymbol z^h_i \rangle_\mathbb{L} = \frac{1}{\kappa}.
 \end{align}
 So $\eta$ will be $\frac{1}{\sqrt{-\kappa}\lvert \lVert \sum_i c_{ij} \boldsymbol z^h_i \rVert_\mathbb{L} \rvert}$, the proof is completed.
\end{proof}

\paragraph{Remark.}
In fact, the geometric centroid in Theorem \ref{theorem.centroid} is also equivalent to the gyro-midpoint in Poincar\'{e} ball model of hyperbolic space.
\begin{proof}
    Let $\boldsymbol z^h_i = [t^h_i, (\boldsymbol s^h_i)^T]^T$, substitute into Theorem \ref{theorem.centroid}, we have
    \begin{align}
        \boldsymbol z^{h-1}_j &= \frac{1}{\sqrt{-\kappa}} \sum_{i=1}^n \frac{ c_{ij}}{\lvert \lVert \sum_{l=1}^n c_{lj} \boldsymbol z^{h}_l  \rVert_\mathbb{L} \rvert} \boldsymbol z^{h}_i \\
        &=\frac{1}{\sqrt{-\kappa}}\frac{[\sum_i c_{ij}t^h_i, \sum_i c_{ij}\boldsymbol s^h_i]^T}{\sqrt{(\sum_i c_{ij}t^h_i)^2 - \lVert \sum_i c_{ij}\boldsymbol s^h_i\rVert^2_2}}.
        \label{eq.expand}
    \end{align}
    Recall the stereographic projection $\mathcal{S}: \mathbb{L}^{\kappa, d_\mathcal{T}} \rightarrow \mathbb{B}^{\kappa, d_\mathcal{T}}$:
    \begin{align}
        \mathcal{S}([t^h_i, (\boldsymbol s^h_i)^T]) = \frac{\boldsymbol s^h_i}{1 + \sqrt{-\kappa}t^h_i}.
    \end{align}
    Then we apply $\mathcal{S}$ to both sides of Eq. (\ref{eq.expand}) to obtain
    \begin{align}
        \boldsymbol b^{h-1}_j &= \frac{1}{\sqrt{-\kappa}}\frac{\sum_i c_{ij}\boldsymbol s^h_i}{\sqrt{(\sum_i c_{ij}t^h_i)^2 - \lVert \sum_i c_{ij}\boldsymbol s^h_i\rVert^2_2} + \sum_i c_{ij}t^h_i} \\
        &= \frac{1}{\sqrt{-\kappa}}\frac{\frac{\sum_i c_{ij}\boldsymbol s^h_i}{\sum_i c_{ij}t^h_i}}{1 + \sqrt{1 - \frac{\lVert \sum_i c_{ij}\boldsymbol s^h_i\rVert^2_2}{(\sum_i c_{ij}t^h_i)^2}}} \\
        &= \frac{\bar{\boldsymbol b}^h}{1 + \sqrt{1 + \kappa \lVert \bar{\boldsymbol b}^h \rVert^2_2}},
    \end{align}
    where $\boldsymbol b^{h-1}_j = \mathcal{S}(\boldsymbol z^{h-1}_j)$, and $\bar{\boldsymbol b}^h = \frac{1}{\sqrt{-\kappa}}\frac{\sum_i c_{ij}\boldsymbol s^h_i}{\sum_i c_{ij}t^h_i}$.
    Let $\boldsymbol b^h_i = \mathcal{S}(\boldsymbol z^h_i)$, then 
    \begin{align}
        \bar{\boldsymbol b}^h = 2\frac{\sum_i c_{ij}
        \frac{\boldsymbol b^h_i}{1 + \kappa \lVert \boldsymbol b^h_i \rVert^2_2}}{\sum_i c_{ij}\frac{1 - \kappa \lVert \boldsymbol b^h_i \rVert^2_2}{1 + \kappa \lVert \boldsymbol b^h_i \rVert^2_2}}
        =\frac{\sum_i c_{ij} \lambda^\kappa_{\boldsymbol b^h_i} \boldsymbol b^h_i }{\sum_i c_{ij}(\lambda^\kappa_{\boldsymbol b^h_i}-1)},
    \end{align}
    which is the gyro-midpoint in Poincar\'{e} ball model. The proof is completed.
\end{proof}

\section{Technical Details}

\subsection{Notations} 
\label{notationTable}

The mathematical notations are described in Table \ref{notation}.
\begin{table}[h]
  \centering
  \begin{tabular}{|c|c|}
    \hline
    Notation & Description \\
    \hline
    $G$ & Graph  \\
    $\mathcal{V}$ & Graph nodes set \\
   $\mathcal{E}$ & Graph edges set\\
    $\mathbf{X}$ & Node attributes matrix \\
    $\mathbf{A}$ & Graph adjacency matrix \\
    $N$ & Graph node number \\
    $w$ & Edge weight function \\
    $v_i$ & A graph node \\
    $d_i$ & The degree of graph node $v_i$ \\
    $\operatorname{Vol}(\cdot)$ & The volume of graph or subset of graph \\
    $\mathcal{U}$ & A subset of $\mathcal{V}$ \\
    $K$ & Number of clusters \\
    $\mathbb{L}^{\kappa, d}$ & $d$-dimensional Lorentz model with curvature $\kappa$ \\
    $\langle \cdot, \cdot \rangle_\mathbb{L}$ & Lorentz inner product \\
    $\mathbf{R}$ & Riemannian metric tensor \\
    $\boldsymbol x$ & Vector or point on manifold \\
    $d_\mathbb{L}(\cdot, \cdot)$ & Lorentz distance \\
    $\lVert \cdot \rVert_\mathbb{L}$ & Lorentz norm \\
    $\lVert \cdot \rVert$ & Euclidean norm \\
    $H$ & Dimension of structural information \\
    $\mathcal{T}$ & A partitioning tree \\
    $\alpha$ & A Non-root node of $\mathcal{T}$ \\
    $\alpha ^ -$ & The immediate predecessor of $\alpha$ \\
    $\lambda$ & Root node of $\mathcal{T}$ \\
    $T_\alpha$ & The module of $\alpha$, a subset of $\mathcal{V}$ associated with $\alpha$ \\
    $g_\alpha$ & The total weights of graph edges with exactly one endpoint in module $T_\alpha$ \\
    $\mathcal{H}^\mathcal{T}(G)$ & Structural information of $G$ w.r.t. $\mathcal{T}$ \\
    $\mathcal{H}^\mathcal{T}(G;\alpha)$ & Structural information of tree node $\alpha$ \\
    $\mathcal{H}^H(G)$ & $H$-dimensional structural entropy of $G$ \\
    $\mathcal{H}^\mathcal{T}(G;h)$ & Structural information of all tree node at level $h$ \\ 
    $\mathbf{C}^h$ & The level-wise assignment matrix between $h$ and $h-1$ level \\
    $N_h$ & The maximal node numbers in $h$-level of $\mathcal{T}$ \\
    $V^h_k$ & The volume of graph node sets $T_k$ \\
    $E(p_1,...,p_n)$ & Entropy function w.r.t. distribution $(p_1,...,p_n)$ \\
    $\Phi(G)$ & Graph conductance \\
    $\mathbf{Z}$ & Tree node embeddings matrix \\
    $\mathbf{\Theta}$ & Learnable parameters of \texttt{LSEnet} \\
    $\mathcal{H}^{\mathcal{T}_{\operatorname{net}}}(G;\mathbf{Z};\mathbf{\Theta})$ & The differentiable structural information (DSI) \\
    $\mathcal{T}_{\operatorname{net}}$ & The partitioning tree decode from \texttt{LSEnet} \\
    $\mathcal O(\cdot)$  & Equivalent infinitesimal \\
    $O(\cdot)$  & The time complexity \\
    \hline
  \end{tabular}
  \caption{The notation descriptions.}
  \label{notation}
\end{table}

\subsection{Algorithm}
\label{append.alg}
The Algorithm \ref{alg:decode} is the decoding algorithm to recover a partitioning tree from hyperbolic embeddings and level-wise assignment matrices. we perform this process in a Breadth First Search manner. 
As we create a tree node object, we put it into a simple queue. 
Then we search the next level of the tree nodes and create children nodes according to the level-wise assignment matrices and hyperbolic embeddings. Finally, we add these children nodes to the first item in the queue and put them into the queue. 
For convenience, we can add some attributes into the tree node objects, such as node height, node number, and so on.

\begin{algorithm}[htb]
    \caption{Decoding a partitioning tree from hyperbolic embeddings in BFS manner.}
    \label{alg:decode}
        \KwIn{Hyperbolic embeddings $\mathbf{Z}^h=\{ \mathbf{z}^h_1, \mathbf{z}^h_2, ..., \mathbf{z}^h_N\}$ for $h=1,...,H$; The height of tree $H$; The matrix $\mathbf{S}^h$ computed in Equation \ref{eq.s}; List of node attributions $node\_attr$, including $node\_set$, $children$, $height$, $coordinates$; Abstract class $\operatorname{Node}$;}
        
    \SetKwFunction{FMain}{ConstructTree}
    \SetKwFunction{COMP}{DFSComps}
    \SetKwFunction{DFS}{DFS}
    \SetKwProg{Fn}{Function}{:}{}
    \Fn{\FMain{$\mathbf{Z}$, $H$, $node\_attr$}}{
        $h = 0$;
        
        \# First create a root node.
        
        $root = \operatorname{Node}(node\_attr[h])$; 

        \# Create a simple queue.
        
        $que = \operatorname{Queue}()$; 
        
        $que.\operatorname{put}(root)$;
        
        \While{$que$ is not empty}
        {
            $node = que.\operatorname{get}()$;

            $node\_set = node.node\_set$
            
            $k = node.height + 1$;

            \# If in the last level of the tree
            
            \eIf{$h == H$}
                {
                    \For{$i$ in $node\_set$}
                    {
                        $child = \operatorname{Node}(node\_attr[h][i])$

                        $child.coordinates = \mathbf{Z}^h_i$
                        
                        $node.children.\operatorname{append}(child)$
                        }
                
                }
                {
                    \For{$k=1$ to $N_h$}
                    {
                        $L\_child$ = [$i$ in where $S^h_{ik}==1$]
                        
                        \If{$\operatorname{len}(L\_child) > 0$}
                        {
                            $child=\operatorname{Node}(node\_attr[h][L\_child])$

                            $child.coordinates = \mathbf{Z}^h_{[L\_child]}$

                            $node.children.\operatorname{append}(child)$

                            $que.\operatorname{put}(child)$
                        }
                    }
                }
        }
        \textbf{return} $root$;
    }
    \KwOut{\FMain{$\mathbf{Z}$, $H$, $node\_attr$}}
\end{algorithm}

In Algorithm \ref{alg:cluster}, we give the approach to obtain node clusters of  a predefined cluster number from a partitioning tree $\mathcal{T}$. The key idea of this algorithm is that we first search the first level of the tree if the number of nodes is more than the predefined numbers, we merge the nodes that are farthest away from the root node, and if the number of nodes is less than the predefined numbers, we search the next level of the tree, split the nodes that closest to the root. As we perform this iteratively, we will finally get the ideal clustering results. The way we merge or split node sets is that: the node far away from the root tends to contain fewer points, and vice versa.

\begin{algorithm}[htb]
    \caption{Obtaining objective cluster numbers from a partitioning tree.}
    \label{alg:cluster}
    \renewcommand{\algorithmicrequire}{\textbf{Input:}}
    \begin{algorithmic}[1]
        \REQUIRE A partitioning tree $\mathcal{T}$; Tree node embeddings $\mathbf{Z}$ in hyperbolic space; Objective cluster numbers $K$.
        \STATE Let $\lambda$ be the root of the tree.
        \FOR{$u$ in $\mathcal{T}.nodes$ / $\lambda$}
            \STATE Compute distance to $\lambda$, i.e., $d_\mathbb{L}(\boldsymbol{z}_\lambda, \boldsymbol{z}_u)$.
        \ENDFOR
        \STATE Sorted non-root tree nodes by distance to $\lambda$ in ascending order, output a sorted list $L$.
        \STATE Let $h=1$ and count the number $M$ of nodes at height $h$.
        \WHILE{$M > K$}
            \STATE Merge two nodes $u$ and $v$ that are farthest away from root $\lambda$.
            \STATE Compute midpoint $p$ of $u$ and $v$, and compute $d_\mathbb{L}(\boldsymbol{z}_\lambda, \boldsymbol{z}_p)$.
            \STATE Add $p$ into $L$ and sort the list again in ascending order.
            \STATE $M=M-1$.
        \ENDWHILE
        \WHILE{$M < K$}
            \FOR{$v$ in $L$}
                \STATE Let $h=h+1$.
                \STATE Search children nodes of $v$ in $h$-level as sub-level list $S$ and count the number of them as $m$.
                \STATE $M=M+ m - 1$.
                \IF{$M > K$}
                    \STATE Perform merge operation to $S$ the same as it to $L$.
                    \STATE Delete $v$ from $L$, and add $S$ to $L$.
                    \STATE Break the for-loop.
                \ELSIF{$M = K$}
                    \STATE Delete $v$ from $L$, and add $S$ to $L$.
                    \STATE Break the for-loop.
                \ELSE
                    \STATE Delete $v$ from $L$, and add $S$ to $L$.
                \ENDIF
            \ENDFOR
        \ENDWHILE
        \STATE Result set $R={}$
        \FOR{$i=0$ to $K-1$}
            \STATE Get graph node subset $Q$ from $L[i]$.
            \STATE Assign each element of $Q$ a clustering category $i$.
            \STATE Add results into $R$.
        \ENDFOR
        \textbf{Output:} $R$
    \end{algorithmic}
    
\end{algorithm}
\subsection{Datasets \& Baselines}
\label{append.database}
We evaluate our model on a variety of datasets, i.e., KarateClub, FootBall, Cora, Citeseer, Amazon-Photo (AMAP), and a larger Computer. All of them are publicly available. We give the statistics of the datasets in Table 1 as follows.

\begin{table}[h]
  \centering
  \begin{tabular}{|c|c|c|c|c|}
    \hline
    Datasets & \# Nodes & \# Features & \# Edges & \# Classes \\
    \hline
    KarateClub & 34 & 34 & 156 & 4 \\
    FootBall & 115 & 115 & 1226 & 12 \\
    Cora & 2708 & 1433 & 5278 & 7 \\
    Citeseer & 3327 & 3703 & 4552 & 6 \\
    AMAP & 7650 & 745 & 119081 & 8 \\
    AMAC & 13752 & 767 & 245861 & 10\\
    \hline
  \end{tabular}
  \caption{The statistics of the datasets.}
\end{table}

In this paper, we compare with deep graph clustering methods and self-supervised GNNs introduced as follows,

\begin{itemize}
  \item[\textbullet] \textbf{RGC}  \cite{liu2023reinforcement} enables the deep graph clustering algorithms to work without the guidance of the predefined cluster number.

  \item[\textbullet] \textbf{DinkNet} \cite{liu2023dink} optimizes the clustering distribution via the designed dilation and shrink loss functions in an adversarial manner.

  \item[\textbullet] \textbf{Congregate} \cite{sun2023contrastive} approaches geometric graph clustering with the re-weighting contrastive loss in the proposed heterogeneous curvature space.

  \item[\textbullet]  \textbf{GC-Flow} \cite{wang2023gc}  models the representation space by using Gaussian mixture, leading to an anticipated high quality of node clustering.

  \item[\textbullet] \textbf{S$^3$GC} \cite{devvrit2022s3gc}  introduces a salable contrastive loss in which the nodes are clustered by the idea of walktrap, i.e., random walk trends to trap in a cluster.

  \item[\textbullet] \textbf{MCGC} \cite{pan2021multi} learns a new consensus graph by exploring the holistic information among attributes and graphs rather than the initial graph.

  \item[\textbullet] \textbf{DCRN} \cite{liu2021deep} designs the dual correlation reduction strategy to alleviate the representation collapse problem.

  \item[\textbullet] \textbf{Sublime}  \cite{liu2022towards} guides structure optimization by maximizing the agreement between the learned structure and a self-enhanced learning target with contrastive learning.

\item[\textbullet] \textbf{gCooL}  \citep{DBLP:conf/www/LiJT22}    clusters nodes with a refined modularity and jointly train the cluster centroids with a bi-level contrastive loss.

  \item[\textbullet] \textbf{FT-VGAE}  \citep{DBLP:conf/ijcai/MrabahBK22} makes effort to eliminate the feature twist issue in the autoencoder clustering architecture by a solid three-step method.

  \item[\textbullet] \textbf{ProGCL} \cite{xia2021progcl} devises two schemes (ProGCL-weight and ProGCLmix) for further improvement of negatives-based GCL methods.

  \item[\textbullet]  \textbf{MVGRL}  \cite{HassaniA20} contrasts across the multiple views, augmented from the original graph to learn discriminative encodings.

  \item[\textbullet] \textbf{ARGA} \cite{pan2018adversarially}  regulates the graph autoencoder with a novel adversarial mechanism to learn informative node embeddings.

  \item[\textbullet] \textbf{GRACE} \cite{yang2017graph} has multiples nonlinear layers of deep denoise autoencoder with an embedding loss, and is devised to learn the intrinsic distributed representation of sparse noisy contents.

  \item[\textbullet] \textbf{DEC} \cite{xie2016unsupervised} uses stochastic gradient descent (SGD) via backpropagation on a clustering objective to learn the mapping.

  \item[\textbullet] \textbf{VGAE} \cite{kipf2016variational} makess use of latent variables and is capable of learning interpretable latent representations for undirected graphs.
\end{itemize}

\subsection{Implementation Notes}
\label{append.note}

\texttt{LSEnet} is implemented upon PyTorch 2.0 \footnote{https://pytorch.org/}, Geoopt \footnote{https://geoopt.readthedocs.io/en/latest/index.html}, PyG \footnote{https://pytorch-geometric.readthedocs.io/en/latest/} and NetworkX \footnote{https://networkx.org/}.
%to implement our proposed model.
%\paragraph{hyperparameter settings}
The dimension of structural information is a hyperparameter, which is equal to the height of partitioning tree.
The hyperbolic partitioning tree is learned in a $2-$dimensional hyperbolic space of Lorentz model by default.
For the training phase, we use Riemannian Adam \cite{becigneul2018riemannian} and set the learning rate to $0.003$.

The loss function of \texttt{LSEnet} is the differentiable structural information (DSI) formulated in Sec. 4.2 of our paper.
We suggest to pretrain $\operatorname{LConv}$ of our model with an auxiliary link prediction loss, given as follows,
% To make the training process stable, we use an auxiliary loss function to pretrain the model. The loss function we choose is the link prediction loss function, formulated as follows:
\begin{align}
    \mathcal{L}_{link} &= \frac{1}{\#(i,j)}\sum_{(i, j)} \mathbf{I}[(i, j) \in \mathcal{E}]\log Pr((i, j) \in \mathcal{E}) + (1 - \mathbf{I}[(i, j) \in \mathcal{E}]) \log [1 - Pr((i, j) \in \mathcal{E})]
\end{align}
where $\#(i, j)$ is the number of the sampling nodes pairs. The probability is defined as 
$Pr((i, j) \in \mathcal{E}) = \frac{1}{1 + \exp{({-\frac{s - d_\mathbb{L}(\boldsymbol z_i, \boldsymbol z_j)}{\tau}}})}$, and $\boldsymbol z_i$ and $\boldsymbol z_j$ are the leaf node embedding of the  hyperbolic partitioning tree.

\section{Additional Results}
\label{append.add}
\begin{figure}[t]
\centering 
\includegraphics[width=1\linewidth]{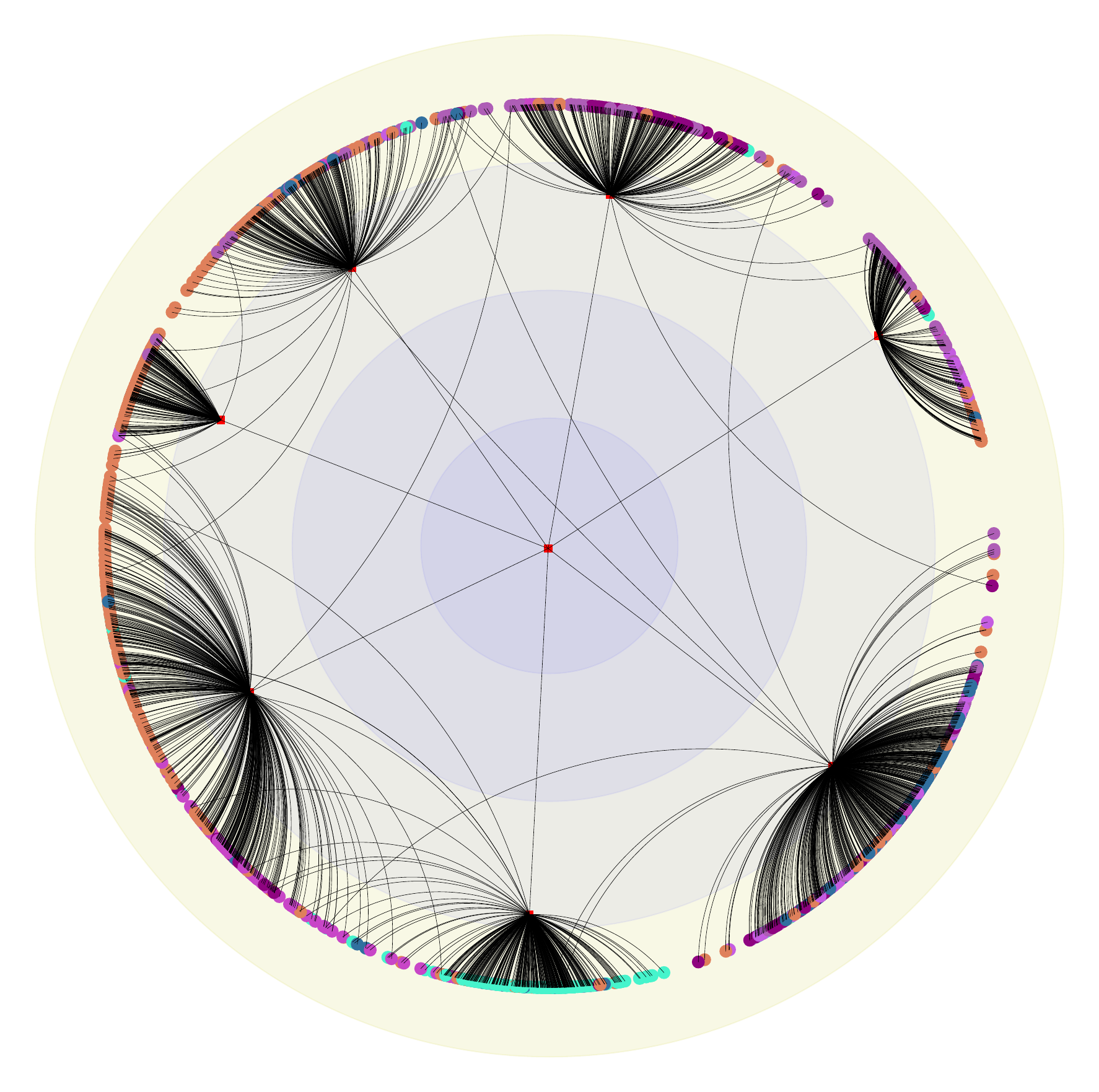}
\caption{Visualization of hyperbolic partitioning trees of Cora.}
\label{Cora.tree}
\end{figure}

The hyperbolic partitioning trees of Cora is visualized in Fig. \ref{Cora.tree}, where different clusters are distinguished by colors.

%%%%%%%%%%%%%%%%%%%%%%%%%%%%%%%%%%%%%%%%%%%%%%%%%%%%%%%%%%%%%%%%%%%%%%%%%%%%%%%
%%%%%%%%%%%%%%%%%%%%%%%%%%%%%%%%%%%%%%%%%%%%%%%%%%%%%%%%%%%%%%%%%%%%%%%%%%%%%%%

\end{document}